%% file: main.tex
\documentclass[lettersize,journal]{IEEEtran}
\usepackage{amsmath,amsfonts}
\usepackage{algorithmic}
\usepackage{array}
\usepackage[caption=false,font=normalsize,labelfont=sf,textfont=sf]{subfig}
\usepackage{textcomp}
\usepackage{url}
\usepackage{verbatim}
\usepackage{graphicx}
 \usepackage{hyperref}
\usepackage{makecell}
\usepackage{amsmath}
\usepackage{amssymb}
\usepackage{mathtools}
\usepackage{amsthm}
\usepackage{graphicx}
\usepackage{float}
\usepackage{dblfloatfix} 
\usepackage{microtype}
\usepackage{graphicx}
\usepackage{subcaption}
\usepackage{booktabs} 
\usepackage{multirow}
\usepackage{adjustbox}
\usepackage{placeins}
\usepackage{xcolor}
\usepackage[ruled,vlined]{algorithm2e}
\usepackage{float}
\usepackage{svg}

\def\BibTeX{{\rm B\kern-.05em{\sc i\kern-.025em b}\kern-.08em
    T\kern-.1667em\lower.7ex\hbox{E}\kern-.125emX}}
\usepackage{balance}

\newcommand{\model}{\textsc{RDDMPI}}

\begin{document}
\title{\model{}: Residual Denoising Diffusion Model for Probabilistic Multivariate Time Series Imputation}
\author{
Ramiro Valdes Jara, 
David Chapman,
Adam Meyers

\thanks{Ramiro Valdes Jara, 
Adam Meyers are with the Department of Industrial and Systems Engineering, University of Miami, United States of America (e-mail: \{rjv71, axm8336\}@miami.edu).}

\thanks{David Chapman is with the Department of Computer Science, University of Miami, United States of America (e-mail: drc201@miami.edu).}

\thanks{Corresponding author: Adam Meyers.}

\thanks{The code can be accessed at \href{https://github.com/ramirovaldesjara/RDDMPI}{https://github.com/ramirovaldesjara/RDDMPI}.}


}


\maketitle

\begin{abstract}
Multivariate time series imputation (MTSI) aims to recover missing values in temporal data composed of multiple interdependent variables. This problem is central to real-world applications such as healthcare monitoring, traffic networks, and energy systems. Recent diffusion-based approaches have shown strong potential for probabilistic imputation by learning to generate missing values through iterative denoising. However, most existing approaches perform diffusion directly in the original data space, requiring the denoising network to simultaneously capture global structure, temporal dynamics, and stochastic variability. This makes the generative task unnecessarily complex, especially when modern deterministic imputers can already provide accurate initial reconstructions. To address this limitation, we propose \model{}, a conditional residual diffusion framework that operates directly in residual space. Instead of modeling the full missing signal directly, we reformulate probabilistic imputation as a baseline-residual decomposition, where a pretrained model captures the dominant signal and a diffusion process models the residual uncertainty. To better exploit deterministic guidance, \model{} conditions the reverse denoising process on both the baseline-completed signal and its latent representation, while a reliability-aware conditioning mechanism adaptively controls the influence of baseline information during residual generation. This formulation simplifies the diffusion learning objective, enabling it to focus on structured correction terms rather than reconstructing the full signal. Experiments on multiple benchmark datasets demonstrate that \model{} consistently improves both reconstruction accuracy and uncertainty quantification. 
\end{abstract}


\input{sections/intro}
\input{sections/relatedwork}

\input{sections/Problem_definition}

\input{sections/methodology}

\input{sections/experiments}

\input{sections/conclusion}

\bibliographystyle{IEEEtran}
\bibliography{references} 

\newpage
\appendices
\onecolumn

\input{sections/appendix}

\end{document}

%% file: sections/intro.tex
\section{Introduction}
\IEEEPARstart{M}{ultivariate} time series data are widely collected in real-world domains such as healthcare monitoring, traffic systems, finance, energy, and climate science \cite{clim, TAN201315, nelwamondo2007missing, HUDAK20082232, JSSv045i03}. In these settings, each variable evolves over time while interacting with other variables. However, missing observations are common due to sensor failures, irregular sampling, communication failures, or data acquisition limitations. Since missing values can degrade downstream tasks such as forecasting, anomaly detection, decision support, and system monitoring, multivariate time series imputation (MTSI) has become a fundamental problem in time series analysis.

The goal of MTSI is to recover missing entries by exploiting both temporal continuity and cross-variable correlations. Early approaches relied primarily on statistical and traditional machine learning techniques, including interpolation methods, matrix and tensor factorization, and Gaussian process models. Although these methods provide interpretable solutions, their ability to capture complex nonlinear temporal dynamics and cross-variable dependencies present in modern multivariate datasets is limited. To overcome these limitations, deep learning-based approaches learn rich temporal and cross-variable representations directly from data, leading to substantial improvements in reconstruction accuracy and transforming the MTSI landscape. 

Existing deep learning approaches can be broadly categorized into deterministic and probabilistic methods. Deterministic methods estimate a single value for each missing entry and have achieved strong performance by capturing temporal dynamics and cross-variable interactions. However, they do not explicitly model uncertainty, which is problematic when multiple plausible imputations are consistent with the observed data. Probabilistic methods address this limitation by modeling a conditional distribution over missing values rather than a single point estimate. Among them, diffusion models have recently emerged as a powerful framework for multivariate time series imputation. By learning an iterative reverse denoising process conditioned on observed entries, diffusion models can generate multiple plausible imputations, thereby enabling uncertainty estimation. 

Despite their success, existing diffusion-based methods typically reconstruct the full missing signal directly from noise. Consequently, the denoising network must jointly learn global temporal structure, cross-variable dependencies, local dynamics, and stochastic variability, resulting in a highly complex learning problem. This observation is particularly relevant because modern deterministic imputers are already capable of recovering much of the underlying signal structure. In many cases, the remaining prediction error contains both systematic, potentially reducible, errors in the reconstruction and irreducible conditional variability arising when multiple missing-value realizations are consistent with the observed data.

This raises a central question: can applying diffusion to residual correction, rather than full-signal reconstruction, simplify the generative learning problem and improve reconstruction accuracy and uncertainty estimation? We argue that residual modeling
leads to a simpler and more targeted generative problem. By
reformulating probabilistic imputation in residual space, diffusion
can focus on correcting systematic baseline errors and modeling the remaining conditional variability rather than reconstructing the full signal from scratch.
However, leveraging deterministic predictions as conditioning information introduces an additional challenge. The quality of baseline imputations can vary across datasets, missingness patterns, and regions of the time series. While accurate baseline predictions provide valuable guidance, unreliable estimates may propagate errors through the diffusion process. Therefore, conditioning information should be incorporated adaptively according to its estimated reliability. 

Motivated by these observations, we propose \model{}, a \textbf{R}esidual \textbf{D}enoising \textbf{D}iffusion \textbf{M}odel for \textbf{P}robabilistic Multivariate Time Series \textbf{I}mputation. RDDMPI decomposes the imputation task into two stages. First, a pretrained deterministic imputation model produces a baseline-completed signal and a structured latent representation. Second, a conditional diffusion model is trained in residual space, where it learns to generate correction terms only for the missing regions. The diffusion process is conditioned on both the baseline reconstruction and its latent representation, allowing the model to focus on probabilistic residual refinement rather than reconstructing the full signal from scratch. In addition, RDDMPI incorporates a reliability-aware conditioning mechanism to adaptively control the influence of baseline information and reduce the propagation of unreliable deterministic estimates.

The main contributions of this work are summarized as follows:
\begin{itemize}
    \item We reformulate probabilistic multivariate time series imputation as a baseline-residual decomposition, separating deterministic signal reconstruction from probabilistic residual uncertainty modeling.
    
    \item We propose RDDMPI, a conditional residual diffusion framework that performs diffusion directly in residual space and leverages both baseline-completed signals and latent baseline representations to guide denoising.
    
    \item We introduce a reliability-aware conditioning mechanism that adaptively modulates the contribution of deterministic baseline information during the reverse diffusion process.
    
    \item We provide extensive empirical evidence across five benchmark datasets showing that RDDMPI achieves state-of-the-art performance in most evaluated settings, consistently outperforming deterministic and probabilistic baselines in both reconstruction accuracy and uncertainty quantification
    
\end{itemize}

The remainder of this paper is organized as follows.
Section~\ref{sec:related_work} reviews existing deterministic
and probabilistic approaches for multivariate time series
imputation. Section~\ref{sec:pre} formulates
the imputation problem. Section~\ref{sec:method} presents
RDDMPI, including the residual diffusion formulation,
the conditional denoising network, the reliability-aware conditioning mechanism, and the theoretical
motivation. Section~\ref{sec:evaluation} reports
experimental results and ablation studies. Finally,
Section~\ref{sec:conclusion} concludes the paper.

%% file: sections/relatedwork.tex
\section{Related Work}
\label{sec:related_work}

Research on multivariate time series imputation (MTSI) has evolved considerably, progressing from deterministic reconstruction methods toward probabilistic generative approaches capable of modeling uncertainty. Existing methods can broadly be categorized into deterministic (Section~\ref{sec:det_tsi}) and probabilistic approaches (Section~\ref{sec:prob_tsi}). In this section, we review these two research directions, with particular emphasis on recent diffusion-based methods that motivate the proposed residual diffusion framework.

\subsection{Deterministic Time Series Imputation}
\label{sec:det_tsi}

Deterministic imputation methods aim to recover missing values by exploiting temporal dependencies and cross-variable correlations present in the observed data. These methods produce a single imputed estimate for each missing entry and have demonstrated strong performance across a wide range of applications.

Traditional approaches include interpolation and smoothing methods~\cite{yi2016stmvl}, as well as low-rank matrix and tensor completion methods~\cite{yu2016trmf,TAN201315}. These methods are efficient and interpretable, exploiting local smoothness or shared low-dimensional structure across variables and time. However, their smoothness and low-rank assumptions limit their ability to capture complex nonlinear dependencies. 

Early deep learning approaches, such as GRU-D \cite{gru_d} and BRITS \cite{brits}, primarily relied on recurrent neural networks (RNNs) to model temporal dynamics while explicitly handling missing observations. However, sequential processing can accumulate reconstruction errors and limit parallelization across timesteps during training and inference.  More recently, attention-based architectures have become increasingly popular due to their ability to capture long-range temporal dependencies and complex interactions among variables. SAITS \cite{saits} employs diagonally masked self attention and combines two reconstruction stages, whereas NRTSI \cite{nrtsi} treats observations as permutation-equivariant sets and progressively imputes missing values without recurrent processing. Other methods jointly model temporal and feature-level dependencies through multidimensional, global-local, or structured attention mechanisms \cite{deepmvi,glima}. 

Recent methods can be divided into two broad categories. One
adapts general-purpose time series architectures, and the other develops specialized architectures for time series imputation.
General-purpose architectures based on temporal convolutions
\cite{moderntcndonghao2024}, two-dimensional temporal variation modeling
\cite{wu2023timesnet}, inverted attention
\cite{liu2024itransformer}, and multiscale mixing
\cite{wang2023timemixer} have been adapted to imputation through
masked reconstruction objectives. These approaches benefit
from advances in general time series representation learning but
are not designed exclusively for missing-value reconstruction.

In contrast, specialized imputation architectures introduce
mechanisms tailored to partially observed data. ImputeFormer
\cite{imputeformer} employs low-rankness-induced attention to
exploit the latent low-dimensional structure of spatiotemporal
data and improve generalization across sensors and missingness
patterns. T1~\cite{tpark2026} introduces one-to-one channel-head
binding, assigning each variable to a dedicated attention head to
strengthen the modeling of variable-specific dynamics while
retaining cross-variable interactions. Beyond architectural design, another line of research incorporates
explicit inductive biases about signal structure or distributional
similarity. Decomposition-based methods separate trend, periodic,
and local components to facilitate the reconstruction of
long-range and recurring patterns~\cite{liu2023multivariate}.
Optimal-transport approaches instead recover missing values by
aligning distributions of time series patches. In particular,
PSW-I~\cite{wang2025optimalpsw-i} incorporates spectral information
into a Wasserstein-based discrepancy, enabling nonparametric
imputation under temporal and distributional shifts without
training a separate predictive model.

For spatiotemporal data, relationships among variables can also
be encoded by a graph. Graph-recurrent architectures
propagate information across both time and connected variables,
while sparse spatiotemporal attention improves scalability to
larger sensor networks~\cite{cini2022filling,marisca2022learning}.
These methods are effective when a meaningful spatial or
relational graph is available. However, general multivariate time
series may not provide such a graph, requiring relationships among
variables to be learned directly from the observed data.

Despite their strong reconstruction performance, deterministic methods inherently produce only a single imputed trajectory. Consequently, they do not explicitly represent the conditional variability of missing values, which can be problematic when multiple plausible completions exist for a partially observed sequence. This limitation has motivated the development of probabilistic approaches that model distributions over missing values rather than single-point estimates.

\subsection{Probabilistic Time Series Imputation}
\label{sec:prob_tsi}

Probabilistic imputation methods seek to estimate the conditional distribution of missing values given the observed data. Unlike deterministic approaches, these methods provide uncertainty estimates alongside imputations, allowing them to represent multiple plausible reconstructions and better characterize ambiguity in the missing regions.

Early probabilistic approaches primarily relied on latent-variable models and Bayesian formulations. Methods such as V-RIN~\cite{v-rin} combine recurrent architectures with variational inference to learn uncertainty-aware latent representations, while Gaussian process-based approaches, including Multi-task GP~\cite{multitask_gp} and GP-VAE~\cite{gp-vae}, provide principled uncertainty quantification through probabilistic priors. However, these methods struggle to capture highly complex temporal dependencies.

More recently, diffusion models have emerged as a powerful probabilistic framework for time series imputation. By learning to reverse a gradual noising process, diffusion models can approximate complex conditional distributions without strong distributional assumptions. CSDI~\cite{csdi} established the foundation for diffusion-based
time series imputation by formulating the task as a conditional
score-based diffusion process, demonstrating that diffusion
models can generate multiple plausible imputations while
providing uncertainty estimates.

Subsequent work has improved diffusion-based imputation through more informative structural conditioning, consistency constraints, and specialized denoising architectures. Spatiotemporal diffusion models incorporate graph and geographic relationships into the conditioning process \cite{Pristi}, while consistency-aware formulations encourage coherent estimates across variables, timesteps, or complementary views \cite{MIDM,mtsci}. FGTI \cite{yang2024frequencyaware} introduces spectral information to better reconstruct periodic and high-frequency components. SSSD
\cite{lopezalcaraz2023diffusionbased} replaces conventional
denoising backbones with structured state-space layers designed
to capture long-range temporal dependencies.

Beyond diffusion, flow matching has recently emerged as an
alternative generative formulation for probabilistic imputation.
GiFlow~\cite{zhang2026spatiotemporal} constructs a graph-informed source
distribution from the observed signal and learns a continuous
transport process toward the target distribution, enabling
efficient generation while explicitly modeling spatial and
temporal dependencies. However, it is specifically designed for
settings in which meaningful graph structure is available.

Despite their differences, existing diffusion-based imputation
methods share a common characteristic: diffusion is performed directly in the original data space. Consequently, the denoising network must learn both the dominant signal structure and the residual uncertainty simultaneously while reconstructing the entire missing signal from noise. This motivates the exploration of alternative formulations that simplify the generative task while retaining the uncertainty modeling benefits of diffusion-based imputation.

\subsection{Residual Modeling and Position of This Work}
Residual modeling has been widely adopted in machine learning as a strategy for simplifying complex prediction tasks.  For example, gradient boosting incrementally improves an initial predictor by fitting successive models to its remaining errors~\cite{friedman2001greedy}. More recently, residual formulations have been incorporated into diffusion models. RDDM~\cite{liu2024residualdenoisingdiffusionmodels} separates image restoration into a directional residual diffusion process and a stochastic noise diffusion process, distinguishing restoration from sample diversity. This idea has recently gained attention in time series forecasting, where RDIT~\cite{lai2025rditresidualbaseddiffusionimplicit} combines a point forecaster with conditional diffusion over its residuals, allowing the deterministic model to capture the central forecast while diffusion models the remaining predictive distribution.

\model{} proposes a residual diffusion formulation for probabilistic multivariate time series imputation. Unlike RDDM, which addresses image restoration, and RDIT, which predicts future values, RDDMPI models residuals only at missing positions. A pretrained deterministic imputer first reconstructs the dominant signal structure, after which diffusion models the conditional distribution of its remaining errors. The denoising process is conditioned on the baseline-completed signal, its latent representation, and the observation mask, while reliability-aware conditioning adaptively controls the influence of potentially inaccurate baseline information. Thus, RDDMPI applies residual diffusion specifically to missing-value reconstruction without requiring the diffusion model to generate the complete missing signal from noise.

%% file: sections/Problem_definition.tex
\section{Problem Definition}
\label{sec:pre}

In this section, we present the mathematical setup and problem definition. Let $X_0 \in \mathbb{R}^{V \times L}$ denote a multivariate
time series with $V$ variables and $L$ timesteps, and let
$M \in \{0,1\}^{V \times L}$ be the corresponding observation
mask, where

\[
M_{v,\ell} =
\begin{cases}
1, & \text{if } X_{0,v,\ell} \text{ is observed},\\
0, & \text{otherwise}.
\end{cases}
\]

Using the mask, the observed and missing components of the
time series can be written as
\begin{gather}
X_0^{\mathrm{ob}} = M \odot X_0,
\\
X_0^{\mathrm{mi}} = (1-M)\odot X_0,
\end{gather}
where $\odot$ denotes element-wise multiplication. The objective of multivariate time series imputation is to
recover the missing component $X_0^{\mathrm{mi}}$ from the
available observations $X_0^{\mathrm{ob}}$.
Formally, an imputation model seeks to learn a mapping
\begin{equation}
\hat{X}_0 = f(X_0^{\mathrm{ob}}, M),
\end{equation}
where $\hat{X}_0$ denotes the reconstructed complete time
series. From a probabilistic perspective, the goal is to estimate the
conditional distribution
\begin{equation}
p(X_0^{\mathrm{mi}} \mid X_0^{\mathrm{ob}}, M),
\end{equation}
which characterizes all plausible values of the missing entries
given the observed context. Deterministic methods approximate
this distribution through a single point estimate, whereas
probabilistic approaches seek to model the full conditional
distribution and provide uncertainty-aware imputations. \model{}
belongs to the latter category and models this conditional
distribution through a residual diffusion process.

%% file: sections/methodology.tex
\section{Methodology}
\label{sec:method}

In this section, we present \model{}, a residual diffusion
framework for probabilistic multivariate time series imputation.
We begin in Section~\ref{sec:residual_decomposition} by
introducing a baseline-residual decomposition that reformulates
imputation as residual modeling around a deterministic
reconstruction. Section~\ref{sec:conditional_residual_diffusion}
then presents a conditional diffusion process that learns the
distribution of residual corrections over missing entries.
Next, Section~\ref{sec:conditional_denoising_network}
describes the proposed conditional denoising network,
including the reliability-aware conditioning mechanism used to
incorporate baseline information during denoising. Finally,
Section~\ref{sec:theory} provides a theoretical motivation for
performing diffusion in residual space rather than directly in
the original data space.
\begin{figure*}[!t]
    \centering
    \includegraphics[width=\textwidth]{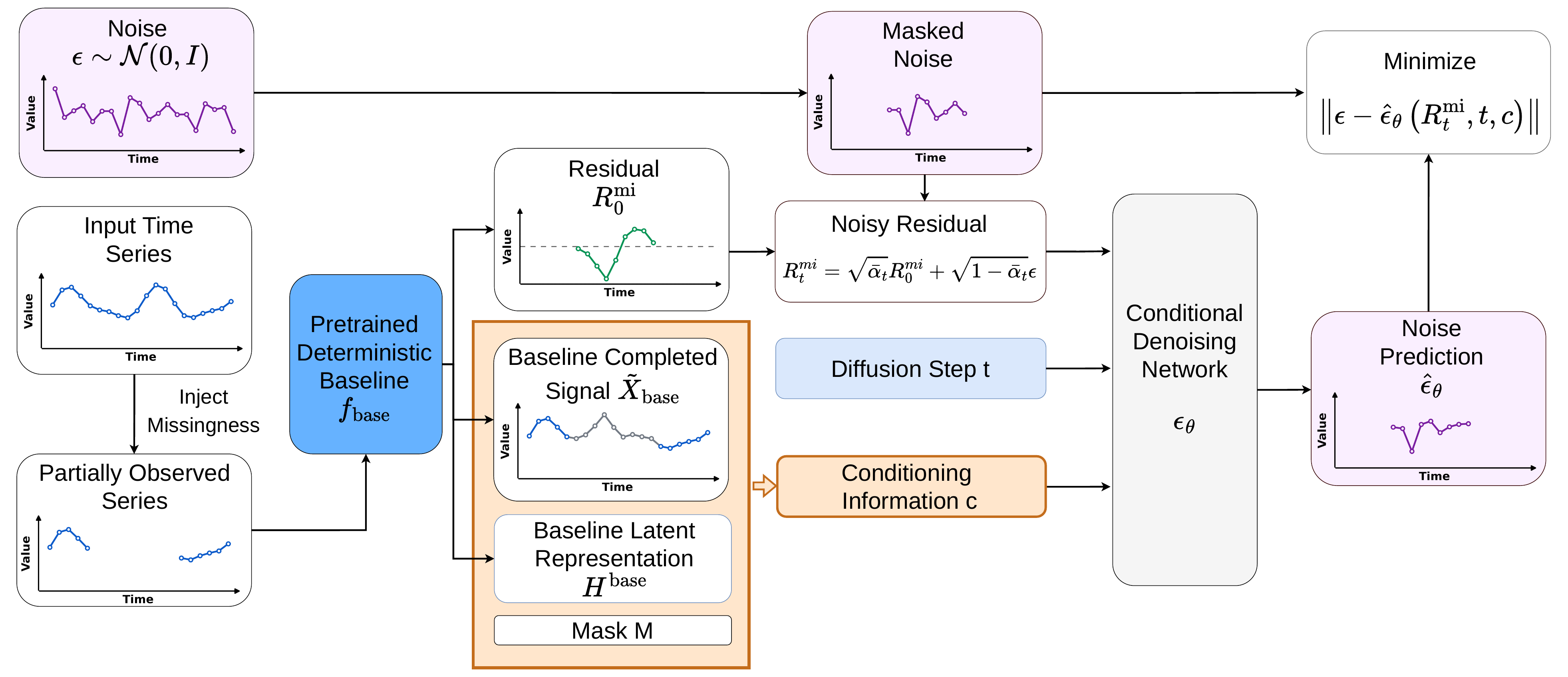}
    \caption{Overview of the residual diffusion training process. Missingness is injected into the input time series to obtain a partially observed series, which is processed by a pretrained baseline to produce a baseline-completed signal, a latent representation and a residual target. Noise is added only to the missing-region residual, and the diffusion model is conditioned on the baseline-completed signal, the baseline latent representation, the noise timestep, and the mask to predict the masked noise. Training minimizes the discrepancy between the true masked noise and the predicted noise.}
    \label{fig:residual_diffusion_training}
\end{figure*}

\subsection{Baseline-Residual Decomposition}
\label{sec:residual_decomposition}

Given the observed component $X_0^{\mathrm{ob}}$ and mask
$M$, a pretrained deterministic imputation model first produces
a baseline-completed signal $\tilde{X}^{\mathrm{base}}$, obtained by filling the missing entries with the baseline model's deterministic predictions, and its latent representation $H^{\mathrm{base}}$ as
\begin{equation}
\tilde{X}^{\mathrm{base}}, H^{\mathrm{base}}
=
f_{\mathrm{base}}(X_0^{\mathrm{ob}}, M).
\end{equation}

Instead of modeling the missing signal directly, we define the
missing-region residual target as
\begin{equation}
R_0^{\mathrm{mi}}
=
(1-M)\odot
\left(
X_0-\tilde{X}^{\mathrm{base}}
\right).
\end{equation}
The residual target is defined only at missing positions, ensuring
that the diffusion model learns corrections to the deterministic
baseline exclusively within the unobserved region. These
corrections capture both systematic baseline errors and the
remaining conditional variability. The proposed residual diffusion model is conditioned on the baseline-completed signal, the baseline latent representation, and the observation mask. We collectively denote this conditioning information as
\begin{equation}
c =
\left(
\tilde{X}^{\mathrm{base}},
H^{\mathrm{base}},
M
\right).
\label{eq:cond_info}
\end{equation}

After sampling a residual correction $\hat{R}_0^{\mathrm{mi}}$, the final imputed
series is obtained by preserving observed entries and correcting
the baseline only on missing positions as
\begin{equation}
\hat{X}_0
=
M\odot X_0
+
(1-M)\odot
\left(
\tilde{X}^{\mathrm{base}}
+
\hat{R}_0^{\mathrm{mi}}
\right).
\end{equation}
Figure~\ref{fig:residual_diffusion_training} illustrates the overall
RDDMPI training process.

\subsection{Conditional Residual Diffusion}
\label{sec:conditional_residual_diffusion}
To model uncertainty over missing values, RDDMPI employs a
conditional Denoising Diffusion Probabilistic Model (DDPM) \cite{ddpm} operating in residual space. DDPMs are latent-variable generative models that learn a data
distribution $p(X_0)$ through a sequence of latent variables
$\{X_t\}_{t=1}^{T}$. These latent variables correspond to
progressively noisier versions of the original data, where
$X_0$ denotes a clean sample and $X_T$ approaches an
isotropic Gaussian distribution after repeated perturbations.
The diffusion framework consists of two Markov processes: a
forward diffusion process that gradually corrupts data by
adding Gaussian noise and a reverse diffusion process that
learns to recover clean samples from noisy observations.

In the proposed framework, we adapt this formulation to the
residual imputation setting by performing diffusion on the
missing-region residual
$R_0^{\mathrm{mi}}$. Let $\{R_t^{\mathrm{mi}}\}_{t=1}^{T}$ denote the noisy residual
states generated by the forward diffusion process. Given the conditioning information
$c$  defined in Eq.~\eqref{eq:cond_info}, the
objective is to learn the conditional residual distribution \(p(R_0^{\mathrm{mi}} \mid c)\), which characterizes plausible residual corrections around the
deterministic baseline reconstruction.
\subsubsection{Forward Diffusion Process}
The forward diffusion process progressively perturbs the clean
residual target with Gaussian noise over $T$ diffusion steps.
Formally, the forward Markov chain is defined as
\begin{gather}
q(R_{1:T}^{\mathrm{mi}} \mid R_0^{\mathrm{mi}})
=
\prod_{t=1}^{T}
q(R_t^{\mathrm{mi}} \mid R_{t-1}^{\mathrm{mi}}),
\\
q(R_t^{\mathrm{mi}} \mid R_{t-1}^{\mathrm{mi}})
=
\mathcal{N}
\left(
R_t^{\mathrm{mi}};
\sqrt{\alpha_t}\,R_{t-1}^{\mathrm{mi}},
(1-\alpha_t)I
\right), \label{eq:forward_transition}
\end{gather}
At each diffusion step, a small amount of Gaussian noise is
added while preserving part of the original residual signal.
As the process progresses, the residual state becomes
increasingly corrupted and gradually loses its structure.
The distribution at
an arbitrary diffusion step $t$ can be written in closed form as
\begin{gather}
q(R_t^{\mathrm{mi}} \mid R_0^{\mathrm{mi}})
=
\mathcal{N}
\left(
R_t^{\mathrm{mi}};
\sqrt{\bar{\alpha}_t}\,R_0^{\mathrm{mi}},
(1-\bar{\alpha}_t)I
\right),
\\
\alpha_t := 1-\beta_t,
\qquad
\bar{\alpha}_t=\prod_{s=1}^{t}\alpha_s.
\end{gather}
where $\beta_t$ denotes the variance schedule and
$\bar{\alpha}_t$ represents the cumulative signal retention
coefficient up to diffusion step $t$. Using the
reparameterization property of DDPMs, a noisy residual at any
diffusion step can be sampled directly as

\begin{equation}
R_t^{\mathrm{mi}}
=
\sqrt{\bar{\alpha}_t}R_0^{\mathrm{mi}}
+
\sqrt{1-\bar{\alpha}_t}\epsilon,
\quad
\epsilon \sim \mathcal{N}(0,I).
\label{eq:forward_noise_injection}
\end{equation}

As $t$ increases, the residual state converges toward a Gaussian noise
distribution. A complete derivation of the residual forward process is provided in Appendix~\ref{app:residual_diffusion_derivation}.

\subsubsection{Reverse Diffusion Process}
The objective of the reverse process is to progressively remove noise from $R_t^{mi}$ while conditioning on $c$. Starting from
$R_T^{\mathrm{mi}}\sim\mathcal{N}(0,I)$, the model learns a
conditional reverse Markov chain
\begin{equation}
\begin{gathered}
p_\theta(R_{0:T}^{\mathrm{mi}} \mid c)
=
p(R_T^{\mathrm{mi}})
\prod_{t=1}^{T}
p_\theta
\left(
R_{t-1}^{\mathrm{mi}}
\mid
R_t^{\mathrm{mi}},t,c
\right),
\\
p_\theta
\left(
R_{t-1}^{\mathrm{mi}}
\mid
R_t^{\mathrm{mi}},t,c
\right)
=
\mathcal{N}
\left(
R_{t-1}^{\mathrm{mi}};
\mu_\theta(R_t^{\mathrm{mi}},t,c),
\sigma_t^2 I
\right).
\end{gathered}
\end{equation}

\noindent where $\mu_\theta(R_t^{\mathrm{mi}},t,c)$ denotes the learned conditional mean of $R_{t-1}^{\mathrm{mi}}$ under the reverse transition, and $\sigma_t^2$ is the corresponding reverse-process variance, fixed according
to the diffusion noise schedule. We set
$\sigma_t^2=\tilde{\beta}_t$, where
$\tilde{\beta}_t =
\frac{1-\bar{\alpha}_{t-1}}{1-\bar{\alpha}_t}\beta_t$. 
Following the standard DDPM parameterization, the reverse mean is expressed through
the noise prediction network $\epsilon_\theta$, whose output
$\hat{\epsilon}_\theta =
\epsilon_\theta(R_t^{\mathrm{mi}},t,c)$ denotes the predicted noise, as
\begin{equation}
\mu_\theta(R_t^{\mathrm{mi}},t,c)
=
\frac{1}{\sqrt{\alpha_t}}
\left(
R_t^{\mathrm{mi}}
-
\frac{1-\alpha_t}
{\sqrt{1-\bar{\alpha}_t}}
\hat{\epsilon}_\theta
\right).
\end{equation}

\subsubsection{Training Objective}
The reverse process introduced above depends on
the noise prediction network
$\epsilon_\theta$,
which is trained to estimate the injected
Gaussian noise $\epsilon$ used to construct $R_t^{mi}$ at an arbitrary diffusion step $t$ in Eq.~\eqref{eq:forward_noise_injection} by minimizing
\begin{equation}
\mathcal{L}_{\theta}
=
\mathbb{E}_{R_t^{\mathrm{mi}},\,\epsilon,\,t}
\left[
\left\|
\left(
\epsilon
-
\hat{\epsilon}_{\theta}
\left(R_t^{\mathrm{mi}},t,c\right)
\right)
\odot
\left(1-M\right)
\right\|_2^2
\right],
\label{eq:residual_loss}
\end{equation}
Importantly, the sampled noise $\epsilon$ is not provided directly as an input to the denoising network. Instead, it serves as the supervision target in the training objective, while $\epsilon_\theta$ receives the noisy residual
$R_t^{\mathrm{mi}}$, the diffusion step $t$, and the conditioning
information $c$. The mask restricts optimization to missing entries,
preventing the model from being penalized on
observed values already fixed by the conditioning
signal. The complete self-supervised training procedure is provided in Appendix~\ref{appendix:train}.

\begin{algorithm}[h]
\small
\SetAlgoLined
\SetKwInOut{Input}{Input}
\SetKwInOut{Output}{Output}

\Input{Partially observed sample $X_0$; mask $M$; pretrained deterministic baseline $f_{\mathrm{base}}$; trained denoising network $\epsilon_\theta$; number of stochastic samples $N$.}

\Output{Imputed sequence $\hat{X}_0$.}

\BlankLine
Compute observed context: $X^{\mathrm{ob}}_0 \leftarrow M \odot X_0$\;

Obtain deterministic baseline-completed signal and latent representation:
$\tilde{X}^{\mathrm{base}},\, H^{\mathrm{base}} \leftarrow f_{\mathrm{base}}(X^{\mathrm{ob}}_0, M)$\;

Define conditioning information:
$c \leftarrow (\tilde{X}^{base},H^{base},M)$;

\For{$n = 1$ \KwTo $N$}{
    Initialize missing-region residual with Gaussian noise:
    $R^{\mathrm{mi},(n)}_T \sim \mathcal{N}(0,I)$\;

    \For{$t = T$ \KwTo $1$}{
        Predict noise using the conditional residual diffusion model:
        $\hat{\epsilon}_\theta
        \leftarrow
        \epsilon_\theta(R_t^{\mathrm{mi},n},t,c)$;

        Compute DDPM reverse mean:
        $\mu_\theta
        \leftarrow
        \frac{1}{\sqrt{\alpha_t}}
        \left(
        R_t^{\mathrm{mi},n}
        -
        \frac{1-\alpha_t}{\sqrt{1-\bar{\alpha}_t}}
        \hat{\epsilon}_\theta
        \right);$\

        Sample reverse step:
        $R^{\mathrm{mi},n}_{t-1} \leftarrow \mu_\theta + \sigma_t z\ , \quad z \sim \mathcal{N}(0,I)$;
    }
}

Estimate final residual correction using the element-wise median:
$\hat{R}^{\mathrm{mi}}_0 \leftarrow
\operatorname{median}\left(\left\{R^{\mathrm{mi},n}_0\right\}_{n=1}^{N}\right)$\;

Recover the missing signal by adding the residual correction to the baseline:
$\hat{X}^{\mathrm{mi}}_0 \leftarrow
(1-M)\odot\left(\tilde{X}^{\mathrm{base}} + \hat{R}^{\mathrm{mi}}_0\right)$\;

Return the full imputed sequence:
$\hat{X}_0 \leftarrow X^{\mathrm{ob}}_0 + \hat{X}^{\mathrm{mi}}_0$\;

\caption{Imputation (Sampling) with \model{}.}
\label{alg:rddmpi_inference}
\end{algorithm}

\subsubsection{Inference}

After training, the denoising network $\epsilon_\theta$
parameterizes the reverse diffusion process. The overall inference procedure of \model{} is summarized in Algorithm~\ref{alg:rddmpi_inference}. The baseline model first provides a deterministic baseline reconstruction and latent representation. Then,
the residual diffusion model performs $N$ independent reverse
sampling trajectories from $t=T$ to $t=1$, producing a set of
residual samples
$\{R^{\mathrm{mi},n}_0\}_{n=1}^{N}$. These samples approximate the conditional residual distribution around the deterministic baseline. Each
sample produces a plausible imputation by adding the sampled
residual to the baseline at missing positions. The complete
sample set is retained for probabilistic evaluation and uncertainty
quantification, whereas its element-wise median is used as the
point estimate for reconstruction-accuracy evaluation. The
architecture used to parameterize
$\epsilon_\theta(R_t^{\mathrm{mi}},t,c)$ is described next in
Section~\ref{sec:conditional_denoising_network}.

\begin{figure*}[!t]
    \centering
    \includegraphics[width=\textwidth]{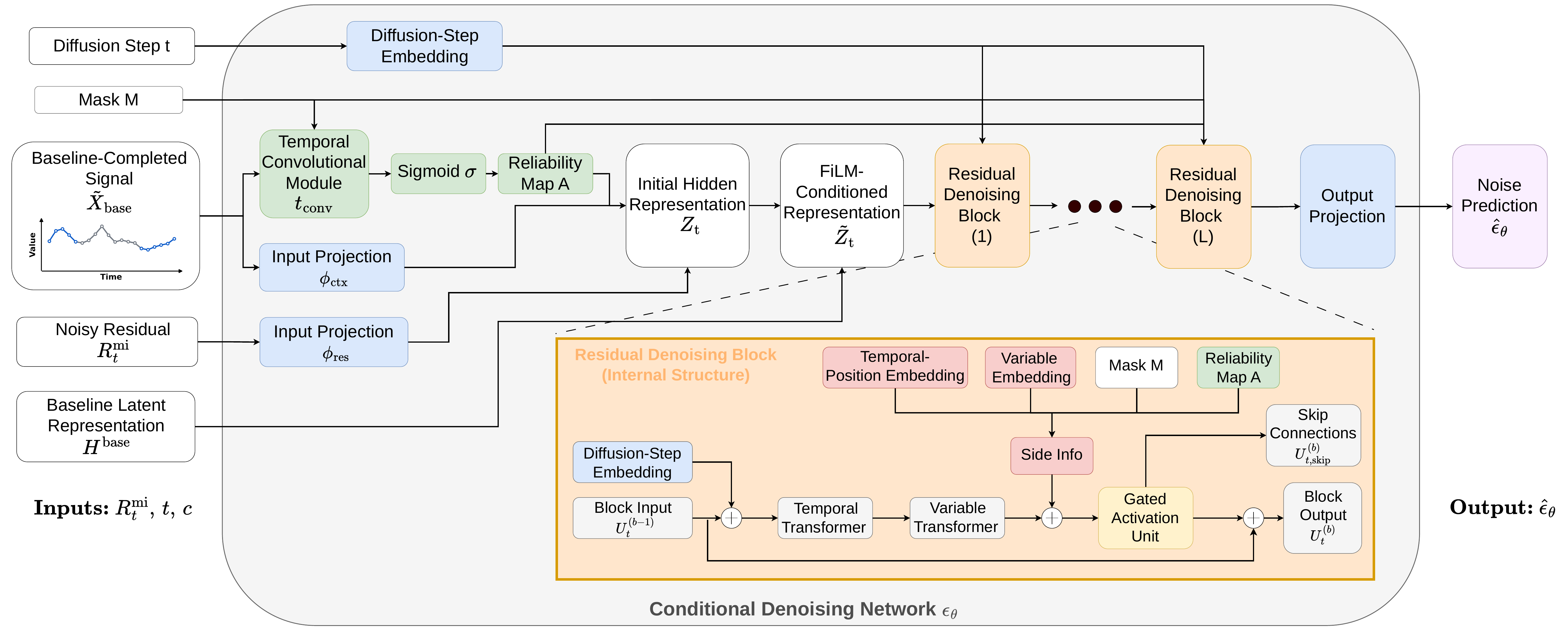}
    \caption{Architecture of the conditional denoising network
    $\boldsymbol{\epsilon}_{\theta}$, which receives the noisy residual
    $R_t^{\mathrm{mi}}$, diffusion step $t$, and conditioning information
    $c=(\widetilde{X}^{\mathrm{base}},H^{\mathrm{base}},M)$, and outputs the noise prediction $\widehat{\epsilon}_{\theta}$. The first residual denoising block receives the FiLM-conditioned
    representation, $U_t^{(0)}=\tilde{Z}_t$, and each subsequent block
    receives the output of the preceding block, $U_t^{(b-1)}$. A complete
    mathematical description of the residual denoising blocks is provided in Appendix~\ref{app:residual_denoising_blocks}.}
    \label{fig:conditional_denoising_network}
\end{figure*}

\subsection{Conditional Denoising Network}
\label{sec:conditional_denoising_network}

The diffusion process is parameterized by a conditional
denoising network $\epsilon_\theta(R_t^{mi},t,c)$.
As illustrated in Figure~\ref{fig:conditional_denoising_network}, the network consists of three main components: a reliability-aware input
fusion mechanism, a FiLM-based conditioning module using the baseline latent representation,
and a stack of residual denoising blocks. Together, these
components allow the denoiser to adaptively incorporate
information from the deterministic baseline while estimating
the noise contained in the current residual state.

\subsubsection{Reliability-Aware Conditioning}

The initial input to the denoising network is the noisy
missing-region residual
$R_t^{\mathrm{mi}}\in
\mathbb{R}^{V\times L}$,
where $V$ is the number of variables
and $L$ is the sequence length. The baseline-completed signal
$\widetilde{X}^{\mathrm{base}}$ is incorporated separately through
a reliability-aware fusion mechanism.

Since the quality of deterministic baseline predictions
can vary across timesteps and variables, conditioning
information should not be incorporated uniformly.
To address this issue, we introduce a learned reliability gate that adaptively controls the influence of the baseline reconstruction on the denoising representation. The reliability
map is computed as
\begin{equation}
A
=
\sigma\!\left(
t_{\mathrm{conv}}\left(
\left[
\tilde{X}^{\mathrm{base}}, M
\right]
\right)
\right).
\label{eq:reliability_gate}
\end{equation}
Here, $t_{\mathrm{conv}}(\cdot)$ is a lightweight temporal convolutional module, $\sigma(\cdot)$ denotes the sigmoid function, and $A\in[0,1]^{V\times L}$ acts as a learned reliability gate over variables and timesteps. The reliability module applies short- and long-range depthwise temporal convolutions independently to each variable, followed by pointwise projections and a sigmoid activation. The resulting map acts as a learned gate rather than an explicitly supervised estimate of baseline error.

The noisy residual and baseline-completed signal are projected
independently to $d_h$ hidden channels. The initial
hidden representation is constructed as
\begin{equation}
Z_t =
\phi_{\mathrm{res}}(R_t^{\mathrm{mi}})
+
A\odot
\phi_{\mathrm{ctx}}
(\widetilde{X}^{\mathrm{base}}),
\end{equation}
where
$\phi_{\mathrm{res}}$ and $\phi_{\mathrm{ctx}}$ are the independent input projections. The reliability map is broadcast across the hidden channels $d_h$.
Thus, the noisy residual defines the primary denoising state,
while the deterministic baseline is incorporated through
reliability-weighted fusion. The resulting representation $Z_t\in\mathbb{R}^{V\times L\times d_h}$ serves as the initial hidden representation, which is subsequently modulated by latent conditioning (Section~\ref{subsec:cond_mecg}) before being processed by the residual denoising blocks. The same reliability map $A$ is passed to later layers as additional conditioning information.

\subsubsection{FiLM-Based Latent Conditioning}
\label{subsec:cond_mecg}
The latent representation extracted from the deterministic
baseline is additionally used to condition the denoising process.
Because its original shape depends on the selected baseline
architecture, it is first aligned with the variable and temporal
dimensions of the denoising representation. A learned projection
then produces multiplicative and additive FiLM \cite{film} tensors as
\begin{equation}
(\gamma_H,\delta_H)
=
g_{\mathrm{FiLM}}
\left(
g_{\mathrm{align}}
\left(H^{\mathrm{base}}\right)
\right),
\end{equation}
where $g_{\mathrm{align}}(\cdot)$ denotes the baseline-specific
reshaping and temporal alignment operation, and
$g_{\mathrm{FiLM}}(\cdot)$ is a learned linear projection.
After dimension permutation,
$\gamma_H,\delta_H\in\mathbb{R}^{V\times L\times d_h}$, FiLM conditioning is
applied as
\begin{equation}
\tilde{Z}_t
=
(1+\gamma_H) \odot Z_t + \delta_H,
\label{eq:film_conditioning}
\end{equation}
where $Z_t$ is the initial hidden representation. This operation injects structured information from the baseline latent representation into the denoising process by adaptively scaling and shifting the hidden features independently across
channels, variables, and timesteps. The result is the FiLM-conditioned hidden
representation
$\tilde{Z}_t\in\mathbb{R}^{V\times L\times d_h}$, which is
provided as input to the residual denoising blocks. Through this
conditioning, the denoiser can exploit the temporal and
cross-variable dependencies encoded by the deterministic
baseline while retaining the noisy residual as its primary
denoising state. 

\subsubsection{Residual Denoising Blocks}

As illustrated in
Figure~\ref{fig:conditional_denoising_network}, the
FiLM-conditioned hidden representation $\widetilde{Z}_t$ is
processed by a stack of residual denoising blocks inspired by
DiffWave~\cite{kong2021diffwave}. We denote the representation
entering the $b$-th block by $U_t^{(b-1)}$, with
$U_t^{(0)}=\widetilde{Z}_t$. Each block produces an updated
representation that is passed to the next block, together with a
skip representation used by the final output path.

Within each block, a projected diffusion-step embedding is first
added to the incoming representation, allowing the denoiser to
adapt to the current noise level. Temporal and variable
transformer layers then process the representation sequentially.
The temporal transformer models dependencies across timesteps
separately for each variable, while the variable transformer
models interactions among variables separately at each timestep.
The resulting features are combined with projected side
information consisting of temporal-position embeddings, variable
embeddings, the observation mask, and the learned reliability map
$A$. 

The conditioned representation is passed through a gated activation
and projected into a residual update and a skip representation.
The residual update is combined with the block input and forwarded
to the next block, while the skip representations from all blocks
are aggregated by the output path. Two final pointwise projections,
with a ReLU activation between them, produce
$
\widehat{\epsilon}_{\theta}
=
\epsilon_{\theta}(R_t^{\mathrm{mi}},t,c)
\in\mathbb{R}^{V\times L},
$
which has the same variable and temporal dimensions as
$R_t^{\mathrm{mi}}$ and represents the Gaussian noise predicted
for the reverse diffusion process. A complete mathematical
description of the block operations, residual and skip
connections, and output aggregation is provided in
Appendix~\ref{app:residual_denoising_blocks}.

\subsection{Theoretical Motivation for Residual Diffusion}
\label{sec:theory}
A natural question is whether performing diffusion in residual space provides a principled advantage over directly modeling the missing signal. Let $X_0^{\text{mi}}$ denote the missing target and let $f(c)$ be a deterministic baseline prediction computed from conditioning information $c$. We define the residual variable on the missing entries as
\begin{equation}
R_0^{\text{mi}} = (1-M) \odot
\left( X_0^{\text{mi}} - f(c)\right).
\end{equation}
This transformation corresponds to a conditional reparameterization rather than a fundamentally different inference problem. In particular, the conditional distributions satisfy
\begin{equation}
p_R(r \mid c) = p_X\big(r + f(c) \mid c\big),
\end{equation}
which implies that residual diffusion is equivalent to standard conditional diffusion under a change of variables.

The practical advantage of this formulation arises when the deterministic baseline removes a substantial fraction of the conditional mean of the missing signal. Let $m(c) = \mathbb{E}[X_0^{\text{mi}} \mid c]$ denote the conditional mean. By adding and subtracting $m(c)$, the residual can be decomposed as
\begin{equation}
R_0^{\text{mi}} = \big(X_0^{\text{mi}} - m(c)\big) + \big(m(c) - f(c)\big).
\end{equation}
Taking squared norms and expectations yields
\begin{align}
\mathbb{E}\!\left[\left\|R^{\mathrm{mi}}_0\right\|^2\right]
={}&
\mathbb{E}\!\left[
\left\|X^{\mathrm{mi}}_0-m(c)\right\|^2
\right]
+
\mathbb{E}\!\left[
\left\|m(c)-f(c)\right\|^2
\right] \notag
\\
&+
\underbrace{
2\mathbb{E}\!\left[
\left\langle
X^{\mathrm{mi}}_0-m(c),
m(c)-f(c)
\right\rangle
\right]
}_{=\,0},
\label{eq:residual_decomposition}
\end{align}
where the cross term vanishes due to conditional centering, since $\mathbb{E}[X_0^{\text{mi}} - m(c)\mid c] = 0$. A complete derivation is provided in Appendix~\ref{app:residual_diffusion_theory_residual}. This decomposition shows that the residual consists of two components: the intrinsic conditional irreducible uncertainty and the squared bias of the deterministic baseline model. Consequently, when $f(c)$ closely approximates $m(c)$, the diffusion model only needs to learn the remaining residual variation rather than the conditional mean structure underlying the missing signal.

This interpretation is particularly relevant in score-based diffusion. Let $s_X(x_t, c, t) = \nabla_{x_t} \log p(x_t \mid c)$ denote the conditional score in data space. When diffusion is performed in residual space, the practical score induced by the baseline, denoted $s_f$, is given by
\begin{equation}
s_f(r_t, c, t) = s_X\big(r_t + \sqrt{\bar{\alpha}_t} f(c), c, t\big),
\end{equation}
whereas the ideal score centered at the conditional mean, denoted $s_m$, is
\begin{equation}
s_m(r_t, c, t) = s_X\big(r_t + \sqrt{\bar{\alpha}_t} m(c), c, t\big).
\end{equation}
Under a standard regularity assumption that $s_X(\cdot, c, t)$ is $L_t$-Lipschitz in its first argument, we obtain
\begin{equation}
\|s_f(r_t, c, t) - s_m(r_t, c, t)\|
\leq
L_t \sqrt{\bar{\alpha}_t} \, \|f(c) - m(c)\|.
\end{equation}
Therefore, if the deterministic baseline accurately approximates the conditional mean, the induced residual score remains close to the ideal centered score. This yields a lower-energy and more correction-oriented target distribution, resulting in a more favorable learning problem for finite-capacity denoising networks in practice. Complete derivations of the score-based analysis are provided in Appendix~\ref{app:residual_diffusion_theory_score}.

%% file: sections/experiments.tex
\section{Evaluation}
\label{sec:evaluation}

In this section, we evaluate RDDMPI on multivariate time
series imputation under different missingness scenarios. We
first describe the datasets, evaluation protocol, baseline
methods, metrics, and implementation details. We then report
quantitative results for reconstruction accuracy and uncertainty
quantification, followed by ablation studies.

\subsection{Evaluation Setup}
\label{sec:evaluation_setup}

\subsubsection{Datasets}

We evaluate RDDMPI on five multivariate time
series benchmark datasets: ETTh1 \cite{zhou_informer_2021}, ETTh2 \cite{zhou_informer_2021}, Weather \cite{wetterstation_weather}, Exchange \cite{lai_2018}, and Illness \cite{cdc_illness}. These
datasets cover diverse domains, including energy, climate,
finance, and public health. For all datasets, we use an input
window length of $96$.

ETTh1 and ETTh2 contain hourly electricity transformer
measurements, each with $7$ correlated variables related to oil
temperature and load covariates. Weather contains $21$
meteorological variables collected at 10-minute intervals from
the Max Planck Institute weather station. Exchange contains
$8$ daily international exchange rate series, representing a
non-stationary financial time series setting. Illness contains
$7$ weekly influenza-related variables from the U.S. Centers for Disease Control and Prevention (CDC) surveillance system. During testing, non-overlapping windows
are used to avoid repeated evaluation over the same temporal
regions.

\subsubsection{Experimental Design}

All experiments are conducted using five random seeds:
$2$, $102$, $202$, $302$, and $402$. Training is performed on
an NVIDIA RTX 6000 GPU. To evaluate generalization under
different observability regimes, we consider both point-wise and structured block missingness.

In the \emph{point missing} setting, we test the model under
missing ratios of $0.2$, $0.4$, $0.6$, and $0.8$, corresponding
to $20\%$, $40\%$, $60\%$, and $80\%$ missing entries, with missing positions sampled independently and uniformly at random. In the
\emph{block missing} setting, we simulate realistic sensor
failure patterns by combining two types of corruption: (i) a
point missing component with rate $0.05$, where $5\%$ of
entries are randomly removed, and (ii) a sequence missing
component with rate $0.0015$, meaning that approximately
$0.15\%$ of positions initiate a missing block. For every initiated block, its length is sampled uniformly from the integers between 24 and 96, and the corresponding consecutive entries of that variable are masked. Blocks extending beyond the end of a window are truncated, and overlapping blocks are combined. This mixed protocol produces
localized measurement dropouts together with extended
contiguous gaps, yielding a more challenging and practically
relevant evaluation scenario.

\subsubsection{Baselines} 

We compare the proposed method against ten representative baselines spanning three model categories: general-purpose time series models, specialized deterministic imputation methods, and specialized probabilistic imputation methods. This distinction is important because some baselines were originally developed for general time series analysis and are adapted here using masked reconstruction, whereas others were designed explicitly for imputation. We briefly summarize the role of each baseline below.

The general-purpose time series models include:
\begin{itemize}
    \item \textbf{DLinear} \cite{DLinear}: A linear baseline that decomposes temporal dynamics through linear projections, providing a strong low-complexity benchmark.
    
    \item \textbf{ModernTCN} \cite{moderntcndonghao2024}: A modern convolutional architecture based on large-kernel temporal convolutions, designed to capture multi-scale temporal patterns efficiently.
    
    \item \textbf{iTransformer} \cite{liu2024itransformer}: A Transformer variant that models multivariate time series through variable-wise tokenization, allowing for capturing cross-variable interactions.
    
    \item \textbf{TimesNet} \cite{wu2023timesnet}: A temporal convolutional model that captures multi-period temporal variation by transforming time series into structured 2D representations.
\end{itemize}

The specialized deterministic imputation methods include:
\begin{itemize}
    \item \textbf{SAITS} \cite{saits}: A self-attention-based imputation model specifically designed for multivariate time series, with masked training objectives for missing-value reconstruction.
    
    \item \textbf{ImputeFormer} \cite{imputeformer}: A transformer-based imputation architecture tailored for generalizable spatiotemporal imputation through low-rank structured attention.

    \item \textbf{T1} \cite{tpark2026}: A CNN-transformer hybrid imputation model that combines temporal convolutional feature extraction with selective cross-variable information transfer through channel-head binding. 
    
\end{itemize}

The probabilistic imputation methods include:
\begin{itemize}
    \item \textbf{GP-VAE}~\cite{gp-vae}: A  latent-variable model that combines a variational autoencoder with a Gaussian-process prior to capture temporal correlations.

    \item \textbf{CSDI} \cite{csdi}: A conditional score-based diffusion model for probabilistic time series imputation, representing a strong diffusion-based baseline with uncertainty-aware generation.

    \item \textbf{FGTI}~\cite{yang2024frequencyaware}: A
    frequency-aware diffusion model that extracts spectral
    information from the observed values to guide the denoising
    process.

\end{itemize}

All baseline implementations are based on established frameworks including Time-Series Library\footnote{\url{https://github.com/thuml/Time-Series-Library}}, PyPOTS ~\cite{du2023pypots}, and Awesome-Imputation ~\cite{du2024tsibench} repositories to ensure reproducibility and fair comparison.

\begin{table*}[!t]
\centering
\caption{Imputation performance on five benchmark datasets under
point and block missing scenarios. Results are averaged across four
point missing ratios (0.2, 0.4, 0.6, 0.8). Dataset abbreviations:
ETTh1/2 = Eh1/2, Exchange = Exch, Illness = Illn, and Weather =
Wthr. Best results are marked in
\textcolor{red}{\textbf{bold}} and second-best results in
\textcolor{blue}{\underline{underlined}}.}
\vspace{-1.5ex}
\label{tab:main_mae_mse}

\adjustbox{width=\textwidth}{%
\setlength{\tabcolsep}{3pt}
\begin{tabular}{l|l|*{10}{cc|}cc}
\toprule

\multicolumn{2}{c|}{Models}
& \multicolumn{2}{c|}{\textbf{\model{} (Ours)}}
& \multicolumn{2}{c|}{DLinear}
& \multicolumn{2}{c|}{ModernTCN}
& \multicolumn{2}{c|}{iTransformer}
& \multicolumn{2}{c|}{SAITS}
& \multicolumn{2}{c|}{ImputeFormer}
& \multicolumn{2}{c|}{TimesNet}
& \multicolumn{2}{c|}{T1}
& \multicolumn{2}{c|}{GP-VAE}
& \multicolumn{2}{c|}{CSDI}
& \multicolumn{2}{c}{FGTI} \\

\multicolumn{2}{c|}{Metric}
& MSE & MAE
& MSE & MAE
& MSE & MAE
& MSE & MAE
& MSE & MAE
& MSE & MAE
& MSE & MAE
& MSE & MAE
& MSE & MAE
& MSE & MAE
& MSE & MAE \\
\midrule

\multirow{2}{*}{\rotatebox{90}{Eh1}}
& Point
& \textcolor{red}{\textbf{0.0548}}
& \textcolor{red}{\textbf{0.1365}}
& 0.2184 & 0.2906
& 0.1145 & 0.2111
& 0.1646 & 0.2569
& 0.1351 & 0.2113
& 0.2856 & 0.3044
& 0.1634 & 0.2567
& 0.0807 & 0.1679
& 0.6246 & 0.5860
& \textcolor{blue}{\underline{0.0641}}
& \textcolor{blue}{\underline{0.1470}}
& 0.0718 & 0.1520 \\

& Block
& \textcolor{red}{\textbf{0.0168}}
& \textcolor{red}{\textbf{0.0851}}
& 0.1728 & 0.2844
& 0.0592 & 0.1735
& 0.1016 & 0.2097
& 0.0257 & 0.1075
& 0.0594 & 0.1543
& 0.0920 & 0.2117
& 0.0258 & 0.1077
& 0.2842 & 0.4231
& 0.0179 & 0.0896
& \textcolor{blue}{\underline{0.0170}}
& \textcolor{blue}{\underline{0.0877}} \\
\midrule

\multirow{2}{*}{\rotatebox{90}{Eh2}}
& Point
& \textcolor{red}{\textbf{0.0442}}
& \textcolor{red}{\textbf{0.1123}}
& 0.0801 & 0.1851
& 0.0576 & 0.1524
& 0.0712 & 0.1745
& 0.4370 & 0.4100
& 0.6667 & 0.4457
& 0.0749 & 0.1799
& \textcolor{blue}{\underline{0.0442}}
& \textcolor{blue}{\underline{0.1264}}
& 1.4139 & 0.8712
& 0.0735 & 0.1444
& 0.0982 & 0.1696 \\

& Block
& \textcolor{red}{\textbf{0.0203}}
& \textcolor{red}{\textbf{0.0744}}
& 0.0770 & 0.1890
& 0.0477 & 0.1414
& 0.0546 & 0.1552
& 0.1471 & 0.2711
& 0.2654 & 0.2672
& 0.0533 & 0.1592
& \textcolor{blue}{\underline{0.0292}}
& \textcolor{blue}{\underline{0.1014}}
& 0.6001 & 0.5663
& 0.0561 & 0.1070
& 0.0948 & 0.1321 \\
\midrule

\multirow{2}{*}{\rotatebox{90}{Exch}}
& Point
& \textcolor{blue}{\underline{0.0022}}
& \textcolor{red}{\textbf{0.0203}}
& 0.0053 & 0.0453
& 0.0095 & 0.0664
& 0.0033 & 0.0339
& 0.2311 & 0.3815
& 0.0693 & 0.1098
& 0.0034 & 0.0335
& \textcolor{red}{\textbf{0.0019}}
& \textcolor{blue}{\underline{0.0218}}
& 0.6415 & 0.6944
& 0.0485 & 0.1135
& 0.0056 & 0.0409 \\

& Block
& 0.0043
& \textcolor{red}{\textbf{0.0181}}
& 0.0063 & 0.0557
& 0.0054 & 0.0498
& \textcolor{red}{\textbf{0.0034}} & 0.0336
& 0.1864 & 0.3330
& 0.1217 & 0.1233
& \textcolor{blue}{\underline{0.0036}} & 0.0362
& 0.0115
& \textcolor{blue}{\underline{0.0312}}
& 0.5100 & 0.6290
& 0.2237 & 0.1458
& 0.0177 & 0.0457 \\
\midrule

\multirow{2}{*}{\rotatebox{90}{Illn}}
& Point
& \textcolor{red}{\textbf{0.0236}}
& \textcolor{red}{\textbf{0.0714}}
& 0.2175 & 0.3046
& 0.0835 & 0.1736
& 0.1211 & 0.2201
& 0.2873 & 0.3044
& 0.3394 & 0.3423
& 0.0969 & 0.2054
& \textcolor{blue}{\underline{0.0252}}
& \textcolor{blue}{\underline{0.0849}}
& 0.6355 & 0.5177
& 0.0806 & 0.1479
& 0.0860 & 0.1464 \\

& Block
& \textcolor{red}{\textbf{0.0390}}
& \textcolor{red}{\textbf{0.1282}}
& 0.2603 & 0.3691
& 0.1521 & 0.2647
& 0.3345 & 0.3598
& 0.1475 & 0.2373
& 0.2697 & 0.3100
& 0.1368 & 0.2535
& 0.0874 & 0.1792
& 0.3270 & 0.4122
& 0.0977 & 0.1895
& \textcolor{blue}{\underline{0.0516}}
& \textcolor{blue}{\underline{0.1283}} \\
\midrule

\multirow{2}{*}{\rotatebox{90}{Wthr}}
& Point
& \textcolor{red}{\textbf{0.0310}}
& \textcolor{red}{\textbf{0.0315}}
& 0.0472 & 0.0882
& 0.0425 & 0.0799
& 0.0920 & 0.1452
& 0.0453 & 0.0647
& 0.0475 & 0.0594
& 0.0471 & 0.0903
& 0.0363 & 0.0582
& 0.1768 & 0.2578
& \textcolor{blue}{\underline{0.0321}}
& \textcolor{blue}{\underline{0.0316}}
& 0.0335 & 0.0320 \\

& Block
& \textcolor{red}{\textbf{0.0233}}
& \textcolor{red}{\textbf{0.0254}}
& 0.0495 & 0.1053
& 0.0371 & 0.0827
& 0.1003 & 0.1480
& 0.0251 & 0.0328
& 0.0401 & 0.0461
& 0.0390 & 0.0855
& 0.0248 & 0.0400
& 0.0627 & 0.1349
& \textcolor{blue}{\underline{0.0234}}
& \textcolor{blue}{\underline{0.0254}}
& 0.0236 & 0.0256 \\

\bottomrule
\end{tabular}%
}
\end{table*}

\subsubsection{Evaluation Metrics}

Let $\mathcal{M}$ denote the set of artificially masked positions used for evaluation, consisting of elements $(v,\ell)$, where $v$ indexes the variable and $\ell$ indexes the timestep. Its cardinality is denoted by $|\mathcal{M}|$. At position $(v,\ell)$, the ground-truth and imputed values are denoted by $y_{v,\ell}$ and $\hat{x}_{v,\ell}$, respectively. For reconstruction accuracy, we use mean absolute error (MAE) and mean squared error (MSE), computed over the positions in $\mathcal{M}$ as
\begin{align}
\mathrm{MAE}
=
\frac{1}{|\mathcal{M}|}
\sum_{(v,\ell)\in\mathcal{M}}
\left|
\hat{x}_{v,\ell}-y_{v,\ell}
\right|,
\end{align}

\begin{align}
\mathrm{MSE}
=
\frac{1}{|\mathcal{M}|}
\sum_{(v,\ell)\in\mathcal{M}}
\left(
\hat{x}_{v,\ell}-y_{v,\ell}
\right)^2.
\end{align}
MAE measures the average reconstruction error, while MSE
penalizes large deviations more strongly. For probabilistic
imputation, we use the continuous ranked probability score (CRPS), which jointly evaluates the accuracy and distributional quality of the predictive distribution. Given a
predictive cumulative distribution function $F_{v,\ell}$, CRPS is
defined as
\begin{align}
\mathrm{CRPS}
=
\frac{1}{|\mathcal{M}|}
\sum_{(v,\ell)\in\mathcal{M}}
\int_{-\infty}^{\infty}
\left(
F_{v,\ell}(z)
-
\mathbb{I}(z\ge y_{v,\ell})
\right)^2 dz.
\end{align}
In this paper, the predictive distribution is approximated using
$N=100$ generated samples
$\{\tilde{x}^{(n)}_{v,\ell}\}_{n=1}^{N}$ at each masked position,
with empirical cumulative distribution
\begin{align}
\hat{F}_{v,\ell}(z)
=
\frac{1}{N}
\sum_{n=1}^{N}
\mathbb{I}
\left(
\tilde{x}^{(n)}_{v,\ell}\le z
\right).
\end{align}
Following the common evaluation protocol for diffusion-based
time series imputation, we compute a normalized quantile-based
approximation. Let
$\mathcal{Q}=\{0.05,0.10,\dots,0.95\}$, and let
$\hat{x}^{(q)}_{v,\ell}$ be the empirical $q$-quantile of the
generated samples. Then,
\begin{align}
\mathrm{CRPS}
=
\frac{1}{|\mathcal{Q}|}
\sum_{q\in\mathcal{Q}}
\frac{
2\sum_{(v,\ell)\in\mathcal{M}}
\rho_q
\left(
y_{v,\ell}-\hat{x}^{(q)}_{v,\ell}
\right)
}{
\sum_{(v,\ell)\in\mathcal{M}}
|y_{v,\ell}|
},
\end{align}
\noindent where the quantile loss is
$\rho_q(u)=u(q-\mathbb{I}(u<0))$.

\subsubsection{Implementation Details}

All methods are evaluated using identical train, validation,
and test splits under the experimental protocol described
above. For general-purpose time series models DLinear \cite{DLinear}, ModernTCN \cite{moderntcndonghao2024}, iTransformer \cite{liu2024itransformer}, and TimesNet \cite{wu2023timesnet}, we employ a
unified training setup using $0.4$ point-wise masking during
training. Optimization is performed using Adam with learning
rate $10^{-3}$, batch size $16$, and a maximum of $300$
epochs with early stopping.

For the specialized imputation models SAITS \cite{saits}, ImputeFormer \cite{imputeformer}, T1 \cite{tpark2026}, GP-VAE \cite{gp-vae}, CSDI \cite{csdi}, and FGTI \cite{yang2024frequencyaware} we preserve the original training procedures and
hyperparameters reported in their official implementations
whenever available. An analysis of parameter count, training cost, and inference efficiency is provided in Appendix~\ref{app:efficiency}.

RDDMPI adopts T1~\cite{tpark2026} as the pretrained
deterministic backbone, providing both the baseline-completed signal and the latent representation used to condition the residual diffusion
model. The deterministic backbone is
pretrained and subsequently frozen during diffusion training.

For the diffusion component, we employ a DDPM formulation with $T=50$ diffusion steps. The variance schedule increases quadratically between the initial and final noise levels, allocating smaller noise increments to the earlier diffusion steps. Its architectural
hyperparameters are selected separately for each dataset. These
include the learning rate, number of residual denoising blocks, and number of hidden channels, among other hyperparameters. The complete
dataset-specific configurations are reported in
Appendix~\ref{app:rddmpi_hyperparameters}.

During training, a self-supervised masking strategy is adopted,
where a masking ratio is randomly sampled and applied to the
observed entries to construct reconstruction targets. During
inference, $N=100$ residual samples are generated through the
reverse diffusion process. The median prediction is used for
deterministic evaluation, while the complete sample set is used
to compute probabilistic metrics such as CRPS and to construct predictive intervals.

\begin{figure*}[!b]
    \centering
    \includegraphics[width=\textwidth]{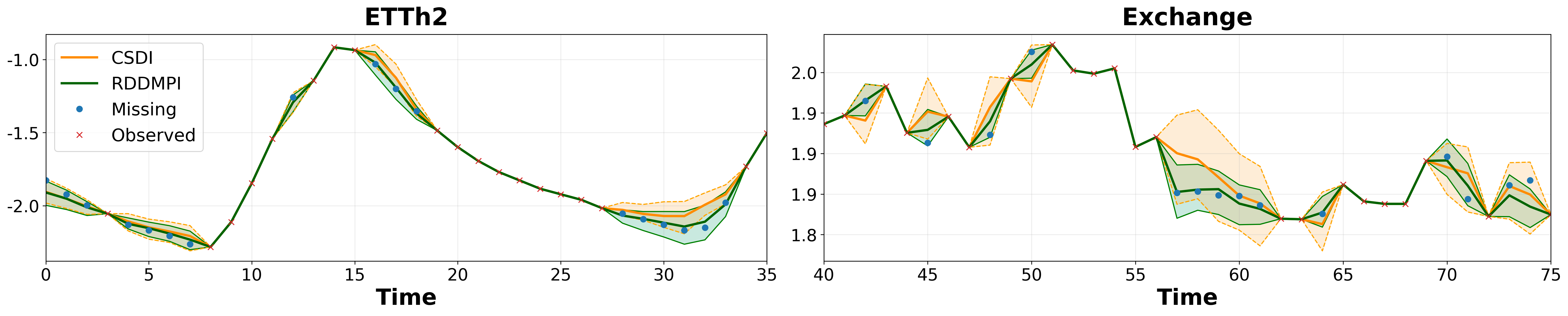}
    \caption{Select examples of probabilistic time series imputation on the ETTh2 and Exchange datasets (zoomed-in views). The red crosses denote observed values, and the blue circles denote the ground-truth imputation targets. For each method, the median imputation is shown as a solid line, while the shaded region represents the 5\% and 95\% predictive quantiles.}
    \label{fig:qualitative_zoom_etth2_exchange}
\end{figure*}

\subsection{Experimental Results}

In this section, we evaluate the proposed method from two complementary perspectives: deterministic reconstruction accuracy and probabilistic uncertainty quantification. Since the goal of probabilistic imputation is not only to recover missing values accurately but also to provide well-calibrated predictive distributions, we report both types of metrics separately for clarity.
The tables in the main paper report results averaged across the four point missing ratios for conciseness. Full results for each individual missing ratio, together with the corresponding standard deviations and block missing results, are provided in Appendix~\ref{appendix:resultsl}.

\subsubsection{Reconstruction Accuracy}

We first assess deterministic imputation quality using MAE and MSE. The corresponding quantitative results across all benchmark datasets are reported in Table~\ref{tab:main_mae_mse}. \model{} achieves the best result in 18 of the 20 comparisons, indicating that residual diffusion provides a robust improvement over both deterministic and probabilistic baselines in reconstruction accuracy. Relative to its deterministic backbone, T1, our proposed model improves in 19 of the 20 comparisons. The improvement is particularly consistent under block missingness, where \model{} outperforms T1 in both MSE and MAE on all five datasets. These results suggest that the benefits of residual diffusion are especially consistent when the deterministic baseline is less accurate.

\subsubsection{Uncertainty Quantification}

We next evaluate probabilistic imputation quality using the continuous ranked probability score (CRPS), which measures the compatibility between the predicted distribution and the observed target values. Since deterministic methods do not model predictive
distributions, CRPS comparisons are reported against GP-VAE, CSDI, and FGTI baselines in Table~\ref{tab:main_crps}.

\begin{table}[t]
\centering
\caption{Probabilistic imputation performance in terms of CRPS
on five benchmark datasets under point and block missing scenarios.
Results for the point missing setting are averaged across four
missing ratios (0.2, 0.4, 0.6, 0.8). Dataset abbreviations:
ETTh1/2 = Eh1/2, Exchange = Exch, Illness = Illn, and Weather =
Wthr. Best results are marked in
\textcolor{red}{\textbf{bold}}, and second-best results are marked in
\textcolor{blue}{\underline{underlined}}.}
\label{tab:main_crps}
\setlength{\tabcolsep}{5pt}

\begin{tabular}{l|l|c|c|c|c}
\toprule
\multicolumn{2}{c|}{Models}
& \textbf{\model{}}
& GP-VAE
& CSDI
& FGTI \\

\multicolumn{2}{c|}{Metric}
& CRPS
& CRPS
& CRPS
& CRPS \\
\midrule

\multirow{2}{*}{\rotatebox{90}{Eh1}}
& Point
& \textcolor{red}{\textbf{0.1313}}
& 0.7376
& \textcolor{blue}{\underline{0.1406}}
& 0.1454 \\

& Block
& \textcolor{red}{\textbf{0.0811}}
& 0.5298
& 0.0867
& \textcolor{blue}{\underline{0.0832}} \\
\midrule

\multirow{2}{*}{\rotatebox{90}{Eh2}}
& Point
& \textcolor{red}{\textbf{0.0641}}
& 0.6395
& \textcolor{blue}{\underline{0.0826}}
& 0.0979 \\

& Block
& \textcolor{red}{\textbf{0.0412}}
& 0.4076
& \textcolor{blue}{\underline{0.0596}}
& 0.0751 \\
\midrule

\multirow{2}{*}{\rotatebox{90}{Exch}}
& Point
& \textcolor{red}{\textbf{0.0179}}
& 0.7848
& 0.0990
& \textcolor{blue}{\underline{0.0354}} \\

& Block
& \textcolor{red}{\textbf{0.0153}}
& 0.6903
& 0.1238
& \textcolor{blue}{\underline{0.0391}} \\
\midrule

\multirow{2}{*}{\rotatebox{90}{Illn}}
& Point
& \textcolor{red}{\textbf{0.0728}}
& 0.6668
& 0.1558
& \textcolor{blue}{\underline{0.1445}} \\

& Block
& \textcolor{blue}{\underline{0.1238}}
& 0.4793
& 0.1873
& \textcolor{red}{\textbf{0.1173}} \\
\midrule

\multirow{2}{*}{\rotatebox{90}{Wthr}}
& Point
& \textcolor{red}{\textbf{0.0420}}
& 0.4501
& \textcolor{blue}{\underline{0.0421}}
& 0.0465 \\

& Block
& \textcolor{blue}{\underline{0.0338}}
& 0.2383
& \textcolor{red}{\textbf{0.0336}}
& 0.0348 \\

\bottomrule
\end{tabular}%
\vspace{-4.5ex}
\end{table}

\model{} obtains the lowest CRPS in eight of the ten dataset and missingness settings. In particular, it obtains the best result on all five
datasets under point missingness and on ETTh1, ETTh2, and Exchange
under block missingness. FGTI performs best on Illness under block
missingness, while CSDI achieves a marginally lower CRPS on
Weather under block missingness.

Taken together, these results show that the advantages of
\model{} are not limited to the median imputation used to compute
MSE and MAE. These results are consistent with the hypothesis that modeling corrections around a deterministic reconstruction provides a more focused generative objective than reconstructing the complete missing signal directly from noise.

Figure~\ref{fig:qualitative_zoom_etth2_exchange} provides a qualitative comparison with CSDI. Both methods recover the local signal morphology, but the median imputations produced by \model{} more closely follow the ground-truth targets, particularly along the descending segment in ETTh2 and the short-term fluctuations
in Exchange. \model{} also produces
narrower predictive intervals that remain centered around the
reconstructed trajectory and contain most of the target values. In these examples, its generated imputations are
therefore less dispersed and maintain close agreement with
the ground truth. Additional qualitative results under different missing
ratios are provided in
Appendix~\ref{app:additional_qualitative_results}.

\begin{table*}[t]
    \centering
    \caption{Component ablation of \model{} on Exchange dataset under point missing and block missing settings. Lower is better.}
    \label{tab:ablation_components}
    \adjustbox{width=\textwidth}{%
    \begin{tabular}{l|ccc|ccc|ccc|ccc|ccc}
    \toprule
    \multirow{3}{*}{Method}
    & \multicolumn{12}{c|}{Point Missing}
    & \multicolumn{3}{c}{Block Missing} \\
    
    & \multicolumn{3}{c|}{0.2}
    & \multicolumn{3}{c|}{0.4}
    & \multicolumn{3}{c|}{0.6}
    & \multicolumn{3}{c|}{0.8}
    & \multicolumn{3}{c}{} \\
    
    & MSE & MAE & CRPS
    & MSE & MAE & CRPS
    & MSE & MAE & CRPS
    & MSE & MAE & CRPS
    & MSE & MAE & CRPS \\
    \midrule

    \makecell[l]{\model{} w/o baseline\\conditioning} 
    & \textcolor{red}{\textbf{0.00184}} 
    & 0.01436 
    & 0.01322 
    & \textcolor{red}{\textbf{0.00129}} 
    & \textcolor{red}{\textbf{0.01536}} 
    & 0.01511 
    & 0.00193 
    & 0.02045 
    & 0.01824 
    & 0.00309 
    & 0.02995 
    & 0.02622 
    & 0.00948 
    & 0.02665 
    & 0.02151 \\

    \midrule

    \makecell[l]{\model{} w/o reliability-\\aware fusion}  
    & \textcolor{red}{\textbf{0.00184}} 
    & \textcolor{red}{\textbf{0.01408}} 
    & \textcolor{red}{\textbf{0.01283}} 
    & \textcolor{red}{\textbf{0.00129}} 
    & 0.01646 
    & \textcolor{red}{\textbf{0.01462}} 
    & \textcolor{red}{\textbf{0.00183}} 
    & \textcolor{red}{\textbf{0.02006}} 
    & \textcolor{red}{\textbf{0.01765}} 
    & 0.00309 
    & 0.02920 
    & 0.02585 
    & 0.01007 
    & 0.03077 
    & 0.02495 \\

    \midrule

    \makecell[l]{\model{}\\(full)}  
    & 0.00210 
    & 0.01465 
    & 0.01312 
    & 0.00148 
    & 0.01705 
    & 0.01511 
    & 0.00193 
    & 0.02055 
    & 0.01814 
    & \textcolor{red}{\textbf{0.00300}} 
    & \textcolor{red}{\textbf{0.02883}} 
    & \textcolor{red}{\textbf{0.02521}} 
    & \textcolor{red}{\textbf{0.00426}} 
    & \textcolor{red}{\textbf{0.01805}} 
    & \textcolor{red}{\textbf{0.01530}} \\
    
    \bottomrule
    \end{tabular}
    }
\end{table*}

\subsection{Ablation Studies}

As a framework, \model{} is designed to extend a strong deterministic imputation baseline with probabilistic residual refinement. We first examine whether its benefits depend on the selected deterministic backbone. We then evaluate the contributions of baseline conditioning and reliability-aware fusion.

\subsubsection{Effect of the Deterministic Backbone}

To determine whether the proposed framework depends on the use
of T1, we instantiate \model{} with either T1 or ImputeFormer as
the pretrained deterministic backbone. Table~\ref{tab:ablation_residual_vs_baseline} compares each complete model with its corresponding deterministic
baseline. RDDMPI reduces both MSE and MAE for both backbones
across every evaluated point and block missingness setting. For
T1, the gains generally increase with the point missing ratio,
indicating that residual correction becomes increasingly useful
as the baseline reconstruction problem becomes more difficult.
The improvements obtained with ImputeFormer are larger because
its baseline errors leave more substantial correction terms for
the residual diffusion model to recover. These results demonstrate that the proposed formulation can generalize to different deterministic imputation architectures, although the
magnitude of the improvement depends on the quality and error
structure of the selected deterministic backbone.

\begin{table}[h]
\centering
\caption{Comparison between deterministic baselines and \model{} on the ETTh1 dataset under point missing and block missing settings. For each deterministic model, \model{} uses that model as its pretrained baseline. IF denotes ImputeFormer. $\Delta$ denotes the percentage improvement of \model{} over its corresponding baseline, where higher $\Delta$ indicates greater improvement. Lower is better for MSE and MAE. Best results are marked in \textcolor{red}{\textbf{bold}}.}
\label{tab:ablation_residual_vs_baseline}
\adjustbox{width=0.99\columnwidth}{%
\setlength{\tabcolsep}{2.5pt}
\begin{tabular}{c l|cc|cc|cc|cc|cc}
\toprule
& \multirow{3}{*}{Method}
& \multicolumn{8}{c|}{Point Missing}
& \multicolumn{2}{c}{Block Missing} \\

& 
& \multicolumn{2}{c|}{0.2}
& \multicolumn{2}{c|}{0.4}
& \multicolumn{2}{c|}{0.6}
& \multicolumn{2}{c|}{0.8}
& \multicolumn{2}{c}{} \\

&
& MSE & MAE
& MSE & MAE
& MSE & MAE
& MSE & MAE
& MSE & MAE \\
\midrule

\multirow{3}{*}{\rotatebox{90}{T1}}
& Base
& 0.0296 & 0.1112
& 0.0397 & 0.1276
& 0.0657 & 0.1632
& 0.1878 & 0.2694
& 0.0258 & 0.1077 \\

& \model{}
& \textcolor{red}{\textbf{0.0228}}
& \textcolor{red}{\textbf{0.0955}}
& \textcolor{red}{\textbf{0.0336}}
& \textcolor{red}{\textbf{0.1122}}
& \textcolor{red}{\textbf{0.0480}}
& \textcolor{red}{\textbf{0.1362}}
& \textcolor{red}{\textbf{0.1144}}
& \textcolor{red}{\textbf{0.2020}}
& \textcolor{red}{\textbf{0.0168}}
& \textcolor{red}{\textbf{0.0850}} \\

\cmidrule(lr){2-12}

& $\Delta$ (\%)
& 22.97 & 14.12
& 15.37 & 12.07
& 26.94 & 16.54
& 39.08 & 25.02
& 34.88 & 21.08 \\

\midrule

\multirow{3}{*}{\rotatebox{90}{IF}}
& Base
& 0.0797 & 0.1697
& 0.1546 & 0.2232
& 0.2931 & 0.3137
& 0.6148 & 0.5108
& 0.0594 & 0.1542 \\

& \model{}
& \textcolor{red}{\textbf{0.0294}}
& \textcolor{red}{\textbf{0.1091}}
& \textcolor{red}{\textbf{0.0402}}
& \textcolor{red}{\textbf{0.1263}}
& \textcolor{red}{\textbf{0.0625}}
& \textcolor{red}{\textbf{0.1566}}
& \textcolor{red}{\textbf{0.1441}}
& \textcolor{red}{\textbf{0.2292}}
& \textcolor{red}{\textbf{0.0208}}
& \textcolor{red}{\textbf{0.0973}} \\

\cmidrule(lr){2-12}

& $\Delta$ (\%)
& 63.11 & 35.71
& 74.00 & 43.41
& 78.68 & 50.08
& 76.56 & 55.13
& 64.98 & 36.90 \\

\bottomrule
\end{tabular}
}
\end{table}

\subsubsection{Component Ablation} To isolate the contributions of the proposed components, we compare three residual diffusion variants on the Exchange dataset that progressively incorporate the proposed design choices.

\textbf{\model{} without baseline conditioning.}
In this variant, the model performs diffusion in residual space but does not receive any conditioning derived from the deterministic backbone. In particular, neither the baseline-completed signal nor the associated latent features are provided to the denoiser. As a result, conditioning mechanisms such as FiLM modulation and reliability-aware fusion are entirely removed. The model therefore relies solely on the noisy residual and diffusion step. This setting isolates the effect of residual diffusion alone.

\textbf{\model{} without reliability-aware fusion.}
Here, baseline conditioning is available, including both the completed signal and latent features from the deterministic backbone. However, the proposed reliability-aware fusion mechanism is removed, meaning that all conditioning information is treated uniformly. This variant corresponds to a conditional residual diffusion formulation and allows us to evaluate whether naive conditioning is sufficient.

\textbf{Full \model{}.}
The full model combines residual diffusion with baseline conditioning and the proposed reliability-aware fusion mechanism. In this setting, the learned reliability gate adaptively modulates the contribution of
baseline-derived information, rather
than incorporating the deterministic context uniformly during denoising.

\textbf{Results analysis.}
Table~\ref{tab:ablation_components} shows that the contributions
of baseline conditioning and reliability-aware fusion depend on
the missingness regime. At point missing ratios 0.2 and 0.4,
the variants without baseline conditioning or without
reliability-aware fusion obtain
the best results. When observations remain sufficiently
distributed throughout the sequence, residual diffusion can
recover the correction term with limited reliance on
baseline-derived guidance.

At point-$0.6$, baseline conditioning without reliability-aware
fusion achieves the best MSE, MAE, and CRPS. This suggests that
the deterministic reconstruction provides useful temporal and
cross-variable guidance at moderate missingness, while its errors
are not yet sufficiently severe to require adaptive gating.
However, this behavior changes in the more difficult regimes.

The complete \model{} obtains the best MSE, MAE, and CRPS at
point-$0.8$ and under block missingness. At point-$0.8$, the
limited number of observed values makes the deterministic
reconstruction less dependable. Under block missingness,
contiguous gaps additionally remove local temporal context.
Incorporating baseline information uniformly can therefore
propagate inaccurate estimates into the denoiser. The
reliability-aware mechanism mitigates this effect by adapting the
contribution of baseline-derived information across variables and
timesteps.

Overall, the ablation results do not indicate that additional
conditioning is uniformly beneficial. Instead, they show that
residual diffusion alone can be sufficient in simpler regimes,
baseline conditioning becomes useful as reconstruction difficulty
increases, and reliability-aware fusion provides its clearest
benefit when baseline errors have a greater effect on the final imputation.

%% file: sections/conclusion.tex
\section{Conclusions}
\label{sec:conclusion}

In this paper, we presented \model{}, a conditional residual diffusion framework for probabilistic multivariate time series imputation. By decomposing the imputation problem into a deterministic initial reconstruction and a residual refinement stage, the proposed approach allows the diffusion model to focus on modeling structured correction terms rather than the full missing signal directly. In addition, the framework leverages both latent representations extracted from the pretrained baseline and reliability-aware conditioning mechanisms to guide denoising under diverse missingness patterns. Across five benchmark datasets, \model{} achieves the best reconstruction result in 18 of the 20 aggregated
MSE and MAE comparisons and the lowest CRPS in eight of the ten
probabilistic comparisons. 

The main limitation of \model{} is its computational cost. The
framework requires a separately pretrained deterministic backbone, and generating $N$ stochastic imputation samples through $T$ reverse steps entails $NT$ denoiser evaluations. Future work will therefore investigate accelerated or reduced-step sampling, and joint training
of the deterministic and diffusion components. Extending the framework to other downstream tasks also represents an important direction for future research.

%% file: sections/appendix.tex
\section{Theoretical Analysis of Residual Diffusion}
\label{app:residual_diffusion_theory}

\subsection{Residual Energy Decomposition}
\label{app:residual_diffusion_theory_residual}
Let $m(c) = \mathbb{E}[X_0^{\mathrm{mi}} \mid c]$ denote the conditional mean. We introduce and subtract $m(c)$ to obtain
\begin{equation}
R_0^{\mathrm{mi}} = \big(X_0^{\mathrm{mi}} - m(c)\big) + \big(m(c) - f(c)\big). \notag
\end{equation} We now analyze the second moment of the residual. Expanding the squared norm yields
\begin{equation}
\|R_0^{\mathrm{mi}}\|^2
=
\|X_0^{\mathrm{mi}} - m(c)\|^2
+
\|m(c) - f(c)\|^2 + 2 \langle X_0^{\mathrm{mi}} - m(c), \, m(c) - f(c) \rangle. \notag
\end{equation}
Taking the conditional expectation with respect to $c$, we obtain
\begin{align}
\mathbb{E}\big[\|R_0^{\mathrm{mi}}\|^2 \mid c\big]
&=
\mathbb{E}\big[\|X_0^{\mathrm{mi}} - m(c)\|^2 \mid c\big] \notag
+
\|m(c) - f(c)\|^2 \nonumber
+
2 \, \mathbb{E}\big[\langle X_0^{\mathrm{mi}} - m(c), \, m(c) - f(c) \rangle \mid c\big]. \notag
\end{align}
\textbf{Vanishing of the cross term.}
We now show that the cross term vanishes. Since $m(c) = \mathbb{E}[X_0^{\mathrm{mi}} \mid c]$, we have
\begin{equation}
\mathbb{E}[X_0^{\mathrm{mi}} - m(c) \mid c] = 0. \notag
\end{equation}
Moreover, $m(c) - f(c)$ is deterministic given $c$. Therefore, using linearity of expectation,
\begin{align}
\mathbb{E}\big[\langle X_0^{\mathrm{mi}} - m(c), \, m(c) - f(c) \rangle \mid c\big] \notag
&=
\left\langle \mathbb{E}[X_0^{\mathrm{mi}} - m(c) \mid c], \, m(c) - f(c) \right\rangle \\ \notag
&= \langle 0, \, m(c) - f(c) \rangle \\ \notag
&=  0.
\end{align}
Thus,
\begin{equation}
\mathbb{E}\big[\|R_0^{\mathrm{mi}}\|^2 \mid c\big]
=
\mathbb{E}\big[\|X_0^{\mathrm{mi}} - m(c)\|^2 \mid c\big]
+
\|m(c) - f(c)\|^2. \notag
\end{equation}
Taking the expectation over $c$ and applying the law of total expectation, we obtain
\begin{equation}
\mathbb{E}\big[\|R_0^{\mathrm{mi}}\|^2\big] \notag
=
\mathbb{E}\big[\|X_0^{\mathrm{mi}} - m(c)\|^2\big]
+
\mathbb{E}\big[\|m(c) - f(c)\|^2\big]. \notag
\end{equation}

\subsection{Score-Based Diffusion Perspective}
\label{app:residual_diffusion_theory_score}
Let $x_t$ denote the noisy variable at diffusion step $t$, and let $p_t(x_t \mid c)$ denote its conditional marginal distribution induced by the forward diffusion process. The corresponding conditional score function is defined as
\begin{equation}
s_X(x_t, c, t) = \nabla_{x_t} \log p_t(x_t \mid c). \notag
\end{equation}
When performing diffusion in residual space, the noisy residual is given by
\begin{equation}
r_t = \sqrt{\bar{\alpha}_t} R_0^{\mathrm{mi}} + \sqrt{1 - \bar{\alpha}_t} \, \epsilon, \quad \epsilon \sim \mathcal{N}(0, I). \notag
\end{equation}
This induces a corresponding noisy variable in data space through the transformation
\begin{equation}
x_t = r_t + \sqrt{\bar{\alpha}_t} f(c). \notag
\end{equation}
The score function associated with this shifted representation is therefore
\begin{equation}
s_f(r_t, c, t) = s_X\big(r_t + \sqrt{\bar{\alpha}_t} f(c), \, c, \, t\big). \notag
\end{equation}

In contrast, the ideal centered residual representation would shift by the conditional mean $m(c) = \mathbb{E}[X_0^{\mathrm{mi}} \mid c]$, leading to the score
\begin{equation}
s_m(r_t, c, t) = s_X\big(r_t + \sqrt{\bar{\alpha}_t} m(c), \, c, \, t\big). \notag
\end{equation}
Assume that for each diffusion step $t$, the score function $s_X(\cdot, c, t)$ is $L_t$-Lipschitz in its first argument, i.e.,
\begin{equation}
\|s_X(u, c, t) - s_X(v, c, t)\| \leq L_t \|u - v\|, \quad \forall u, v. \notag
\end{equation}
Applying the Lipschitz property with
\[
u = r_t + \sqrt{\bar{\alpha}_t} f(c), \quad v = r_t + \sqrt{\bar{\alpha}_t} m(c), \notag
\]
we obtain
\begin{align}
\|s_f(r_t, c, t) - s_m(r_t, c, t)\| =
\big\| s_X(r_t + \sqrt{\bar{\alpha}_t} f(c), c, t) \notag
- s_X(r_t + \sqrt{\bar{\alpha}_t} m(c), c, t) \big\| \leq L_t \sqrt{\bar{\alpha}_t} \|f(c) - m(c)\|. \notag
\end{align}
This inequality shows that, under regularity conditions, the discrepancy between the practical residual score $s_f$ and the ideal centered score $s_m$ is controlled by the baseline approximation error $\|f(c) - m(c)\|$. In particular, if $f(c)$ provides an accurate approximation of the conditional mean $m(c)$, then the induced residual score remains close to the optimal centered score. Moreover, the discrepancy is modulated by the diffusion coefficient $\sqrt{\bar{\alpha}_t}$, implying that the influence of baseline error diminishes as the diffusion process progresses toward higher noise levels. Consequently, residual diffusion yields a score field that is both stable and closer to the ideal centered representation, leading to a simpler and more favorable learning problem for a finite capacity denoising network.

\section{Mathematical Details of Forward Process Derivation of Residual Diffusion}
\label{app:residual_diffusion_derivation}

This appendix derives the closed-form expression for the noisy
residual $R_t^{\mathrm{mi}}$ given in
Eq.~\eqref{eq:forward_noise_injection}, starting from the
one-step forward diffusion transition in Eq.~\eqref{eq:forward_transition}.

The one-step forward transition is
\begin{equation}
R_t^{mi} = \sqrt{\alpha_t} R_{t-1}^{mi} + \sqrt{1-\alpha_t}\epsilon_t,
\quad \epsilon_t \sim \mathcal{N}(0, \mathrm{I}).
\notag
\end{equation}
Similarly,
\begin{equation}
R_{t-1}^{mi} = \sqrt{\alpha_{t-1}} R_{t-2}^{mi} + \sqrt{1-\alpha_{t-1}}\epsilon_{t-1},
\quad \epsilon_{t-1} \sim \mathcal{N}(0, \mathrm{I}).
\label{eq:appendix_forward_previous}
\end{equation}
Substituting Eq.~\eqref{eq:appendix_forward_previous} into the one-step forward transition, we obtain
\begin{equation}
\begin{aligned}
R_t^{mi}
&= \sqrt{\alpha_t} \left(
\sqrt{\alpha_{t-1}} R_{t-2}^{mi}
+ \sqrt{1-\alpha_{t-1}}\epsilon_{t-1}
\right)
+ \sqrt{1-\alpha_t}\epsilon_t \\
&= \sqrt{\alpha_t \alpha_{t-1}} R_{t-2}^{mi}
+ \sqrt{\alpha_t(1-\alpha_{t-1})}\epsilon_{t-1}
+ \sqrt{1-\alpha_t}\epsilon_t.
\end{aligned}
\notag
\end{equation}
Continuing this expansion recursively, we obtain
\begin{align}
R_t^{mi}
&= \sqrt{\alpha_t \alpha_{t-1} \cdots \alpha_1} R_0^{mi} \notag \\
&\quad + \sqrt{\alpha_t \alpha_{t-1} \cdots \alpha_2 (1-\alpha_1)}\epsilon_1 \notag \\
&\quad + \sqrt{\alpha_t \alpha_{t-1} \cdots \alpha_3 (1-\alpha_2)}\epsilon_2 \notag \\
&\quad + \cdots \notag \\
&\quad + \sqrt{\alpha_t(1-\alpha_{t-1})}\epsilon_{t-1}
+ \sqrt{1-\alpha_t}\epsilon_t. \notag
\end{align}
Now define
$
\bar{\alpha}_t = \prod_{i=1}^t \alpha_i.
$ Thus, the first term becomes
$
\sqrt{\bar{\alpha}_t} R_0^{mi}.
$ Now consider the noise terms. Since $\epsilon_i \sim \mathcal{N}(0, \mathrm{I})$ are i.i.d., each scaled term is
\begin{align}
\sqrt{\alpha_t \cdots \alpha_{i+1}(1-\alpha_i)}\epsilon_i
\sim \mathcal{N}\left(0, \alpha_t \cdots \alpha_{i+1}(1-\alpha_i)\mathrm{I}\right). \notag
\end{align}
Since linear combinations of independent Gaussian variables remain Gaussian, the sum of these terms is also Gaussian with variance given by the sum of the individual variances; that is,
\begin{equation}
\sum_{i=1}^t \alpha_t \cdots \alpha_{i+1}(1-\alpha_i)
= 1 - \bar{\alpha}_t. \notag
\end{equation}
Therefore, the total noise term becomes
\begin{equation}
\sqrt{1-\bar{\alpha}_t} \, \epsilon, \quad \epsilon \sim \mathcal{N}(0, \mathrm{I}). \notag
\end{equation}
Finally, we obtain
\begin{equation}
R_t^{mi}
= \sqrt{\bar{\alpha}_t} R_0^{mi}
+ \sqrt{1-\bar{\alpha}_t}\epsilon,
\quad \epsilon \sim \mathcal{N}(0, \mathrm{I}). \notag
\end{equation}
This yields the closed-form forward noising expression used in
Eq.~\eqref{eq:forward_noise_injection}.

\section{\model{} Training Procedure}
\label{appendix:train}

Following the conditional residual diffusion formulation described in Section~\ref{sec:method}, let $X_0 \in \mathbb{R}^{V \times L}$ denote a training window sampled from the data distribution, with $V$ variables and sequence length $L$. We use $v\in\{1,\ldots,V\}$ to index variables and $\ell\in\{1,\ldots,L\}$ to index timesteps. Let $M \in \{0,1\}^{V \times L}$ be the binary observation mask, where $M_{v,\ell}=1$ indicates an observed value, whereas $M_{v,\ell}=0$ indicates a missing value. 
During training, a subset of originally observed entries is randomly masked and treated as missing targets. The residual diffusion process is then applied only to the missing-region residual $R^{\mathrm{mi}}_0$, while the observed mask, the baseline-completed signal, and the latent baseline representation remain fixed and serve as conditioning information. At diffusion step $t$, Gaussian noise $\epsilon \sim \mathcal{N}(0,I)$ is injected into the missing-region residual
\begin{equation}
R^{\mathrm{mi}}_t
=
\sqrt{\bar{\alpha}_t}\,R^{\mathrm{mi}}_0
+
\sqrt{1-\bar{\alpha}_t}\,\epsilon, \notag
\end{equation}
where $\{\alpha_t\}_{t=1}^{T}$ is a predefined variance schedule and
$\bar{\alpha}_t=\prod_{s=1}^{t}\alpha_s$.
The model is trained to predict the injected noise using a conditional noise predictor
\begin{equation}
\epsilon_\theta\!\left(R^{\mathrm{mi}}_t,\, t ,\, c\right). \notag
\end{equation}
Since diffusion is applied only to the missing-region residual, the denoising loss is computed exclusively on missing positions:
\begin{equation}
\mathcal{L}(\theta)
=
\mathbb{E}
\left[
\frac{1}{\sum_{v,\ell}(1-M_{v,\ell})}
\sum_{v=1}^{V}\sum_{\ell=1}^{L}
(1-M_{v,\ell})
\left(
\epsilon_{v,\ell}
-
\epsilon_\theta(\cdot)_{v,\ell}
\right)^2
\right]. \notag
\end{equation}

The self-supervised training procedure of the proposed residual diffusion model is summarized in Algorithm~\ref{alg:rddm_training}.

\begin{algorithm}[H]
\caption{Training of \model{}}
\label{alg:rddm_training}
\begin{algorithmic}[1]
\STATE {\bfseries Input:} training data distribution $q(X_0)$, pretrained deterministic baseline $f_{\mathrm{base}}$, number of optimization iterations $N_{\rm iter}$, noise schedule $\{\alpha_t\}_{t=1}^{T}$ with $\bar{\alpha}_t = \prod_{s=1}^{t}\alpha_s$
\STATE {\bfseries Output:} trained conditional noise predictor $\epsilon_\theta$
\FOR{$i = 1$ {\bfseries to} $N_{\rm iter}$}
    \STATE Sample a training window $X_0 \sim q(X_0)$ and a diffusion timestep $t \sim \mathrm{Uniform}(\{1,\ldots,T\})$
    \STATE Sample a mask by randomly hiding a subset of currently observed entries, producing $M$
    \STATE Compute observed and missing components: $X^{\mathrm{ob}}_0 \leftarrow M \odot X_0,\; X^{\mathrm{mi}}_0 \leftarrow (1-M)\odot X_0$
    \STATE Obtain baseline-completed signal and latent representation: $\tilde{X}^{\mathrm{base}},\, H^{\mathrm{base}} \leftarrow f_{\mathrm{base}}(X^{\mathrm{ob}}_0, M)$
    \STATE Construct the missing-region residual target: $R^{\mathrm{mi}}_0 \leftarrow (1-M)\odot \left(X_0 - \tilde{X}^{\mathrm{base}}\right)$
    \STATE Sample Gaussian noise $\epsilon \sim \mathcal{N}(0,I)$ with the same shape as $R^{\mathrm{mi}}_0$
    \STATE Apply forward noising to the residual target: $R^{\mathrm{mi}}_t \leftarrow \sqrt{\bar{\alpha}_t}\,R^{\mathrm{mi}}_0 + \sqrt{1-\bar{\alpha}_t}\,\epsilon$
    \STATE Predict noise using the residual diffusion model: $\hat{\epsilon} \leftarrow \epsilon_\theta\!\left(R^{\mathrm{mi}}_t,\, t,\, c\right)$
    \STATE Take a gradient step minimizing 
      $
        \nabla_\theta
        \left(
        \frac{1}{\sum_{v,\ell}(1-M_{v,\ell})}
        \sum_{v=1}^{V}\sum_{\ell=1}^{L}
        (1-M_{v,\ell})
        \left(
        \epsilon_{v,\ell} - \hat{\epsilon}_{v,\ell}
        \right)^2
        \right)
    $
\ENDFOR
\end{algorithmic}
\end{algorithm}

\section{Efficiency Analysis}
\label{app:efficiency}

Table~\ref{tab:efficiency} presents computational efficiency and performance metrics across \model{}, DLinear \cite{DLinear}, ModernTCN \cite{moderntcndonghao2024}, iTransformer \cite{liu2024itransformer}, TimesNet \cite{wu2023timesnet}, SAITS \cite{saits}, ImputeFormer \cite{imputeformer}, T1 \cite{tpark2026}, GP-VAE \cite{gp-vae}, CSDI \cite{csdi} and FGTI \cite{yang2024frequencyaware}. 

\begin{table}[H]
\centering
\caption{Computational efficiency and performance comparison on the ETTh1 and Illness datasets. Params (M): parameters in millions; Train Speed: ms per iteration; Inference Speed: ms per sample; MAE: Mean Absolute Error averaged over point missing ratios 0.2, 0.4, 0.6, and 0.8 (lower is better).}
\label{tab:efficiency}

\adjustbox{width=0.85\textwidth}{%
\setlength{\tabcolsep}{3pt}
\begin{tabular}{l|l|c|c|c|c}
\toprule
Dataset & Model & Parameters (M) & Train Speed (ms/iter) & Inference Speed (ms/sample) & MAE \\
\midrule

\multirow{11}{*}{ETTh1}
& \model{} (Ours) & 1.23 & 213.96  & 2220.24 & 0.1365 \\
& DLinear          & 0.02 & 4.02    & 0.16    & 0.2906 \\
& ModernTCN        & 1.72 & 7.42    & 0.50    & 0.2111 \\
& iTransformer     & 0.22 & 7.27    & 0.08    & 0.2569 \\
& TimesNet         & 0.59 & 22.06   & 1.28    & 0.2567 \\
& SAITS            & 5.27 & 51.68   & 0.28    & 0.2113 \\
& ImputeFormer     & 1.37 & 21.83   & 0.47    & 0.3044 \\
& T1               & 0.54 & 12.80   & 0.33    & 0.1679 \\
& GP-VAE           & 0.11 & 23.04   & 19.60   & 0.5860 \\
& CSDI             & 1.19 & 188.98 & 2315.40 & 0.1470 \\
& FGTI             & 1.96 & 45.72   & 3696.13 & 0.1520 \\

\midrule

\multirow{11}{*}{Illness}
& \model{} (Ours) & 0.44 & 313.97 & 12614.79 & 0.0714 \\
& DLinear          & 0.02 & 5.48   & 1.07     & 0.3046 \\
& ModernTCN        & 0.76 & 22.10  & 3.46     & 0.1736 \\
& iTransformer     & 0.22 & 18.56  & 1.70     & 0.2201 \\
& TimesNet         & 4.69 & 52.15  & 15.81    & 0.2054 \\
& SAITS            & 5.27 & 83.61  & 6.32     & 0.3044 \\
& ImputeFormer     & 1.37 & 39.84  & 5.79     & 0.3423 \\
& T1               & 0.54 & 15.24  & 5.47     & 0.0849 \\
& GP-VAE           & 0.11 & 23.19  & 19.24    & 0.5177 \\
& CSDI             & 0.41 & 268.52 & 13997.47 & 0.1479 \\
& FGTI             & 0.52 & 87.67  & 18958.83 & 0.1464 \\

\bottomrule
\end{tabular}%
}

\vspace{-1ex}
\end{table}

Table~\ref{tab:efficiency} shows that \model{} has a substantially higher inference
cost than deterministic baselines, although it achieves better
imputation accuracy on both datasets. This difference is primarily
caused by iterative probabilistic sampling rather than the model size. Deterministic methods produce a point estimate through a single
forward pass, whereas generating one stochastic imputation sample
with a diffusion model requires a complete reverse trajectory of
$T$ sequential denoising steps. At each step, the current noisy
state is updated according to
\begin{equation}
x_{t-1}
=
\frac{1}{\sqrt{\alpha_t}}
\left(
x_t
-
\frac{1-\alpha_t}{\sqrt{1-\bar{\alpha}_t}}
\epsilon_{\theta}(x_t,t,\mathrm{cond})
\right)
+
\sigma_t z,
\qquad
z\sim\mathcal{N}(0,I). \notag
\end{equation}

Also, a single reverse trajectory produces one imputed sample, i.e.,
one possible estimate of the missing values. To approximate the
predictive distribution, \model{} generates $N$ independent
samples, each requiring its own $T$-step reverse trajectory.
Inference therefore requires $NT$ denoising-network evaluations
in total. The resulting samples are aggregated to obtain the
reported point imputation and are also used to quantify predictive
uncertainty. Although the $N$ trajectories may be evaluated in
parallel through batching, the $T$ denoising steps within each
trajectory remain sequential. Consequently, inference cost depends
on both the number of diffusion steps $T$ and the number of generated
samples $N$, explaining the higher inference times of \model{},
CSDI, and FGTI relative to single-pass deterministic methods.

\paragraph{Accelerated Diffusion Sampling.}
To mitigate this limitation, we explore faster sampling strategies such as Denoising Diffusion Implicit Models (DDIM) \cite{song2022denoisingdiffusionimplicitmodels}, which reduce the number of required steps while maintaining performance. DDIM introduces a non-Markovian deterministic sampling process:
\begin{equation}
\mathrm{x}_{t-1} = \sqrt{\bar{\alpha}_{t-1}} \hat{\mathrm{x}}_0 + \sqrt{1 - \bar{\alpha}_{t-1}} \, \epsilon_\theta(\mathrm{x}_t, t), \notag
\end{equation}
where $\hat{\mathrm{x}}_0$ is the predicted clean signal. By selecting a subset of timesteps $\{t_1, \dots, t_K\}$ with $K \ll T$, inference can be significantly accelerated. We evaluate DDIM-based sampling with reduced step counts and observe that \model{} maintains competitive performance even with substantially fewer inference steps, while achieving notable speedups. This highlights the potential of accelerated diffusion methods to bridge the efficiency gap with deterministic models.

\begin{table}[H]
\centering
\caption{Results of \model{} under point missing ratios (0.2, 0.4, 0.6, 0.8). This table compares the default setting ($T=50$) with a reduced number of steps ($T=10$), along with the relative degradation $\Delta$ in \%.}
\label{tab:efficiency_ddim}
\setlength{\tabcolsep}{6pt}
\renewcommand{\arraystretch}{0.95}

\begin{tabular}{l| l|ccc|ccc|ccc|}
\toprule
\multicolumn{2}{c|}{Models} & \multicolumn{3}{c|}{\model{} ($T = 50$)} & \multicolumn{3}{c|}{\model{} ($T = 10$)} & \multicolumn{3}{c|}{$\Delta$ (in \%)}\\
 \multicolumn{2}{c|}{Metric} & MSE & MAE & CRPS & MSE & MAE  & CRPS & MSE & MAE  & CRPS\\
\midrule

\multirow{4}{*}{\rotatebox{90}{ETTh1}}
& 0.2 & 0.0229 & 0.0956 & 0.0924
      & 0.0236 & 0.0970 & 0.0957
      & +3.06 & +1.46 & +3.57 \\

& 0.4 & 0.0337 & 0.1123 & 0.1080
      & 0.0345 & 0.1139 & 0.1117
      & +2.37 & +1.42 & +3.43 \\

& 0.6 & 0.0480 & 0.1362 & 0.1310
      & 0.0495 & 0.1390 & 0.1358
      & +3.13 & +2.06 & +3.66 \\

& 0.8 & 0.1145 & 0.2020 & 0.1940
      & 0.1216 & 0.2102 & 0.2031
      & +6.20 & +4.06 & +4.69 \\
\midrule

\multirow{4}{*}{\rotatebox{90}{ETTh2}}
& 0.2 & 0.0228 & 0.0771 & 0.0438
      & 0.0223 & 0.0786 & 0.0455
      & -2.19 & +1.95 & +3.88 \\

& 0.4 & 0.0305 & 0.0924 & 0.0528
      & 0.0295 & 0.0935 & 0.0540
      & -3.28 & +1.19 & +2.27 \\

& 0.6 & 0.0437 & 0.1150 & 0.0657
      & 0.0422 & 0.1157 & 0.0666
      & -3.43 & +0.61 & +1.37 \\

& 0.8 & 0.0797 & 0.1649 & 0.0941
      & 0.0760 & 0.1646 & 0.0941
      & -4.64 & -0.18 & 0.00 \\

\midrule

\multirow{4}{*}{\rotatebox{90}{Exchange}}
& 0.2 & 0.0022 & 0.0147 & 0.0131
      & 0.0022 & 0.0150 & 0.0138
      & 0.00 & +2.04 & +5.34 \\

& 0.4 & 0.0015 & 0.0171 & 0.0151
      & 0.0042 & 0.0185 & 0.0169
      & +180.00 & +8.19 & +11.92 \\

& 0.6 & 0.0019 & 0.0206 & 0.0181
      & 0.0035 & 0.0219 & 0.0198
      & +84.21 & +6.31 & +9.39 \\

& 0.8 & 0.0030 & 0.0288 & 0.0252
      & 0.0030 & 0.0290 & 0.0257
      & 0.00 & +0.69 & +1.98 \\
\midrule

\multirow{4}{*}{\rotatebox{90}{Illness}}
& 0.2 & 0.0057 & 0.0449 & 0.0460
      & 0.0059 & 0.0468 & 0.0494
      & +3.51 & +4.23 & +7.39 \\

& 0.4 & 0.0081 & 0.0518 & 0.0526
      & 0.0092 & 0.0547 & 0.0563
      & +13.58 & +5.60 & +7.03 \\

& 0.6 & 0.0132 & 0.0664 & 0.0680
      & 0.0151 & 0.0702 & 0.0727
      & +14.39 & +5.72 & +6.91 \\

& 0.8 & 0.0673 & 0.1225 & 0.1248
      & 0.0720 & 0.1355 & 0.1370
      & +6.98 & +10.61 & +9.78 \\
\midrule

\multirow{4}{*}{\rotatebox{90}{Weather}}
& 0.2 & 0.0246 & 0.0246 & 0.0327
      & 0.0233 & 0.0258 & 0.0361
      & -5.28 & +4.88 & +10.40 \\

& 0.4 & 0.0271 & 0.0271 & 0.0362
      & 0.0270 & 0.0289 & 0.0401
      & -0.37 & +6.64 & +10.77 \\

& 0.6 & 0.0313 & 0.0317 & 0.0423
      & 0.0321 & 0.0336 & 0.0462
      & +2.56 & +5.99 & +9.22 \\

& 0.8 & 0.0409 & 0.0425 & 0.0573
      & 0.0412 & 0.0441 & 0.0599
      & +0.73 & +3.76 & +4.54 \\
      
\bottomrule

\end{tabular}

\end{table}

Future work should focus on improving inference efficiency through advanced sampling techniques, such as adaptive timestep selection or one-step diffusion models. These directions could further reduce computational cost while preserving the strong performance benefits of diffusion-based imputation.

\section{Residual Denoising Block Architecture}
\label{app:residual_denoising_blocks}

This appendix provides a detailed description of the residual
denoising blocks used in the conditional noise prediction network
$\epsilon_{\theta}(R_t^{\mathrm{mi}},t,c)$. Let $V$ denote the number of variables,
$L$ the sequence length, $d_h$ the number of hidden channels, and
$N_{\mathrm{blk}}$ the number of residual denoising blocks.

The reliability-aware fusion and FiLM-based conditioning
mechanisms described in Section~\ref{sec:conditional_denoising_network} produce the FiLM-conditioned hidden representation $\tilde{Z}_t$, which is the input to the residual denoising stack and initialized as
$
U_t^{(0)}
=
\tilde{Z}_t
\in
\mathbb{R}^{V\times L\times d_h}.
\notag
$
This representation is passed sequentially through
$N_{\mathrm{blk}}$ residual denoising blocks. For block
$b\in\{1,\ldots,N_{\mathrm{blk}}\}$, we define
\begin{align}
U_t^{(b,0)}
=
U_t^{(b-1)},
\notag
\end{align}
where $U_t^{(b-1)}$ is the output of the preceding block. Each
block then applies diffusion-step conditioning, temporal
attention, variable attention, side-information conditioning, a
gated activation, and residual and skip projections. The complete
sequence of operations is summarized in
Algorithm~\ref{alg:residual_denoising_block} and described below.

\subsection{Diffusion-Step Conditioning}
\label{subsec:diff-step}
The discrete diffusion step $t$ is represented by a fixed
sinusoidal embedding, followed by two learned linear projections
with SiLU activations as
\begin{align}
e_t
=
\operatorname{SiLU}
\left(
W_{e,2}
\operatorname{SiLU}
\left(
W_{e,1}e_t^{(0)}+b_{e,1}
\right)
+b_{e,2}
\right),
\notag
\end{align}
where $e_t^{(0)}$ is the fixed sinusoidal representation, $d_{\mathrm{diff}}$ denotes the dimension of the diffusion-step
embedding, and
$e_t\in\mathbb{R}^{d_{\mathrm{diff}}}$ is the resulting
diffusion-step embedding. Within block $b$, this embedding is projected to the hidden
dimension as
\begin{align}
d_t^{(b)}
=
W_{\mathrm{diff}}^{(b)}e_t
+
b_{\mathrm{diff}}^{(b)}
\in
\mathbb{R}^{d_h}.
\notag
\end{align}
The projected embedding is broadcast across variables and
timesteps and added to the incoming representation:
\begin{align}
U_t^{(b,1)}
=
U_t^{(b,0)}
+
\operatorname{Broadcast}
\left(
d_t^{(b)}
\right).
\notag
\end{align}
Thus,
$U_t^{(b,1)}\in\mathbb{R}^{V\times L\times d_h}$ contains both
the hidden residual features and information about the current
diffusion noise level.

\subsection{Temporal and Cross-Variable Modeling}

The representation is first processed by a temporal transformer
independently for each variable as
\begin{align}
U_t^{(b,2)}[v,:,:]
=
\mathcal{T}_{\mathrm{time}}^{(b)}
\left(
U_t^{(b,1)}[v,:,:]
\right),
\qquad
v=1,\ldots,V.
\notag
\end{align}
For each variable, the corresponding
$L\times d_h$ slice is treated as a temporal sequence. Therefore,
the temporal transformer models dependencies across timesteps
without mixing the variable dimension. The resulting representation is subsequently processed by a
variable transformer independently at each timestep as
\begin{align}
U_t^{(b,3)}[:,\ell,:]
=
\mathcal{T}_{\mathrm{var}}^{(b)}
\left(
U_t^{(b,2)}[:,\ell,:]
\right),
\qquad
\ell=1,\ldots,L.
\notag
\end{align}
At each timestep, the corresponding $V\times d_h$ slice is
treated as a sequence over variables. The variable transformer
therefore models cross-variable interactions while preserving the
temporal positions.

Both transformer modules use multi-head self-attention, residual
connections, layer normalization, and a position-wise
feed-forward network with a GELU activation. For an input
sequence $X$, an individual attention head is computed as
\begin{align}
\operatorname{head}_{j}(X)
=
\operatorname{softmax}
\left(
\frac{
(XW_j^Q)(XW_j^K)^{\top}
}{
\sqrt{d_k}
}
\right)
XW_j^V,
\notag
\end{align}
and the multi-head output is
\begin{align}
\operatorname{MHA}(X)
=
\operatorname{Concat}
\left(
\operatorname{head}_1(X),
\ldots,
\operatorname{head}_H(X)
\right)
W^O,
\notag
\end{align}
where $H$ is the number of attention heads and $d_k$ is the
dimension of each query and key head.

\subsection{Side-Information Conditioning}
Let
$S\in\mathbb{R}^{V\times L\times d_{\mathrm{side}}}$
denote the side-information tensor provided to every residual
denoising block. The side information consists of a
temporal-position embedding, a learned variable embedding, the
observation mask, and the learned reliability map.

For each temporal coordinate $\tau_\ell$, we construct a
$d_{\mathrm{time}}$-dimensional sinusoidal embedding. Its
components are given by
\begin{align}
E_{\mathrm{time}}(\ell,2j)
&=
\sin
\left(
\frac{\tau_\ell}
{10000^{2j/d_{\mathrm{time}}}}
\right),
\notag\\
E_{\mathrm{time}}(\ell,2j+1)
&=
\cos
\left(
\frac{\tau_\ell}
{10000^{2j/d_{\mathrm{time}}}}
\right).
\notag
\end{align}
The same temporal embedding is replicated across all variables,
resulting in
$
E_{\mathrm{time}}
\in
\mathbb{R}^{V\times L\times d_{\mathrm{time}}},
$ where $d_{\mathrm{time}}$ denotes the dimension of the temporal-position embedding.
Each variable $v$ is assigned a learned embedding vector
$e_v\in\mathbb{R}^{d_{\mathrm{var}}}$, where
$d_{\mathrm{var}}$ denotes the dimension of the variable
embedding, through an embedding
lookup table:
$
e_v
=
\operatorname{Emb}_{\mathrm{var}}(v).
\notag
$
This embedding is replicated across all timesteps to obtain
$
E_{\mathrm{var}}
\in
\mathbb{R}^{V\times L\times d_{\mathrm{var}}}.
\notag
$ The observation mask $M\in\{0,1\}^{V\times L}$ indicates which
entries are available as conditioning information. The learned
reliability map $A\in[0,1]^{V\times L}$ is computed by the
reliability-aware mechanism described in Section~\ref{sec:conditional_denoising_network}.
After treating $M$ and $A$ as one-dimensional feature channels,
the complete side-information tensor is
\begin{align}
S
=
\operatorname{Concat}
\left(
E_{\mathrm{time}},
E_{\mathrm{var}},
M,
A
\right)
\in
\mathbb{R}^{V\times L\times d_{\mathrm{side}}},
\notag
\end{align}
where $d_{\mathrm{side}}$ denotes the total dimension of the
side-information features and is given by
$
d_{\mathrm{side}}
=
d_{\mathrm{time}}
+
d_{\mathrm{var}}
+
2,
\notag
$ with the additional two dimensions corresponding to the observation mask $M$ and reliability map $A$.
The reliability map therefore affects the network twice: first
during the reliability-weighted fusion of the baseline-completed
signal and again as side information within every residual
denoising block.

The output of the variable transformer and the side-information
tensor are projected independently to $2d_h$ hidden dimensions.
Let $\psi_{\mathrm{mid}}^{(b)}$ and
$\psi_{\mathrm{cond}}^{(b)}$ denote the learned pointwise
projections in block $b$, implemented as $1\times1$
convolutions. Specifically,
$\psi_{\mathrm{mid}}^{(b)}$ maps $d_h$ to $2d_h$ dimensions,
whereas $\psi_{\mathrm{cond}}^{(b)}$ maps
$d_{\mathrm{side}}$ to $2d_h$ dimensions. Their outputs are
combined as
\begin{align}
U_t^{(b,4)}
=
\psi_{\mathrm{mid}}^{(b)}
\left(
U_t^{(b,3)}
\right)
+
\psi_{\mathrm{cond}}^{(b)}
\left(
S
\right),
\notag
\end{align}
where
$
U_t^{(b,4)}
\in
\mathbb{R}^{V\times L\times 2d_h}.
\notag
$
Because both operators are pointwise, the same learned
transformation is applied independently at every
variable-timestep position, without mixing information across
variables or timesteps.

\subsection{Gated Activation}

The conditioned representation is split equally along its hidden
dimension as
\begin{align}
\left(
U_{t,\mathrm{gate}}^{(b,4)},
U_{t,\mathrm{filter}}^{(b,4)}
\right)
=
\operatorname{Split}
\left(
U_t^{(b,4)}
\right),
\notag
\end{align}
where both components belong to
$\mathbb{R}^{V\times L\times d_h}$. A gated activation is then applied as
\begin{align}
U_t^{(b,5)}
=
\sigma
\left(
U_{t,\mathrm{gate}}^{(b,4)}
\right)
\odot
\tanh
\left(
U_{t,\mathrm{filter}}^{(b,4)}
\right).
\notag
\end{align}
The sigmoid component controls how much information is propagated,
while the hyperbolic-tangent component provides the nonlinear
candidate features. Their element-wise product gives
$U_t^{(b,5)}\in\mathbb{R}^{V\times L\times d_h}$.

\subsection{Residual and Skip Outputs}

The gated representation is projected pointwise from $d_h$ to
$2d_h$ hidden dimensions. Let
$\psi_{\mathrm{out}}^{(b)}$ denote the learned block-output
projection, implemented as a $1\times1$ convolution:
$
\psi_{\mathrm{out}}^{(b)}
:
\mathbb{R}^{V\times L\times d_h}
\rightarrow
\mathbb{R}^{V\times L\times 2d_h}.
\notag
$
The projected representation is therefore
\begin{align}
U_t^{(b,6)}
=
\psi_{\mathrm{out}}^{(b)}
\left(
U_t^{(b,5)}
\right)
\in
\mathbb{R}^{V\times L\times 2d_h}.
\notag
\end{align}
Because the projection is pointwise, it transforms the hidden
features independently at every variable-timestep position,
without mixing information across variables or timesteps.

The resulting representation is divided equally along the hidden
dimension into a residual update and a skip connection as
\begin{align}
\left(
U_{t,\mathrm{res}}^{(b)},
U_{t,\mathrm{skip}}^{(b)}
\right)
=
\operatorname{Split}
\left(
U_t^{(b,6)}
\right),
\notag
\end{align}
where
$
U_{t,\mathrm{res}}^{(b)},
U_{t,\mathrm{skip}}^{(b)}
\in
\mathbb{R}^{V\times L\times d_h}.
\notag
$ The residual output is added to the block input and scaled by
$1/\sqrt{2}$
\begin{align}
U_t^{(b)}
=
\frac{
U_t^{(b,0)}
+
U_{t,\mathrm{res}}^{(b)}
}{
\sqrt{2}
}.
\notag
\end{align}
The resulting tensor $U_t^{(b)}$ becomes the input to the next
residual denoising block. The factor $1/\sqrt{2}$ helps control
the magnitude of the hidden activations across the sequence of
blocks. The skip output $U_{t,\mathrm{skip}}^{(b)}$ is retained
for the final noise prediction path.

\subsection{Skip Aggregation and Noise Prediction}

After all residual denoising blocks have been evaluated, their
skip outputs are summed and normalized as
\begin{align}
U_{t,\mathrm{skip}}
=
\frac{1}{\sqrt{N_{\mathrm{blk}}}}
\sum_{b=1}^{N_{\mathrm{blk}}}
U_{t,\mathrm{skip}}^{(b)}.
\notag
\end{align}

The aggregated skip representation is processed by two learned
pointwise projections. The first projection,
$\psi_{\mathrm{out},1}$, is a $1\times1$ convolution that
preserves the hidden dimension:
$
\psi_{\mathrm{out},1}
:
\mathbb{R}^{V\times L\times d_h}
\rightarrow
\mathbb{R}^{V\times L\times d_h}.
\notag
$ It is followed by a ReLU activation as
\begin{align}
U_t^{\mathrm{out}}
=
\operatorname{ReLU}
\left(
\psi_{\mathrm{out},1}
\left(
U_{t,\mathrm{skip}}
\right)
\right).
\notag
\end{align}

The second projection,
$\psi_{\mathrm{out},2}$, is a $1\times1$ convolution that maps
the $d_h$ hidden features to one predicted noise value at every
variable-timestep position:
$
\psi_{\mathrm{out},2}
:
\mathbb{R}^{V\times L\times d_h}
\rightarrow
\mathbb{R}^{V\times L}.
\notag
$ The final noise prediction is
\begin{align}
\widehat{\epsilon}_{\theta}
=
\psi_{\mathrm{out},2}
\left(
U_t^{\mathrm{out}}
\right)
\in
\mathbb{R}^{V\times L}.
\notag
\end{align}
Both output projections operate independently at each
variable-timestep position. The temporal and cross-variable
interactions have already been incorporated by the transformer
layers within the residual denoising blocks.

Therefore,
$
\widehat{\epsilon}_{\theta}
=
\epsilon_{\theta}
\left(
R_t^{\mathrm{mi}},
t,
c
\right)
\notag
$
has the same variable and temporal dimensions as the noisy
residual and represents the Gaussian noise predicted for the
reverse diffusion process.

\begin{algorithm}[!t]
\caption{Sequential Operations of the Residual Denoising Network}
\label{alg:residual_denoising_block}

\KwIn{FiLM-conditioned representation
$U_t^{(0)}=\widetilde{Z}_t$; diffusion step $t$; side-information
tensor $S$; number of residual blocks $N_{\mathrm{blk}}$}

\KwOut{Predicted noise
$\widehat{\epsilon}_{\theta}
=\epsilon_{\theta}(R_t^{\mathrm{mi}},t,c)$}

Compute the diffusion-step embedding $e_t$ (Subsection~\ref{subsec:diff-step}) \;

Initialize an empty collection of skip representations\;

\For{$b=1$ \KwTo $N_{\mathrm{blk}}$}{
    Set the block input:
    $U_t^{(b,0)}\leftarrow U_t^{(b-1)}$\;

    Project and inject the diffusion-step embedding:
    $U_t^{(b,1)}
    \leftarrow U_t^{(b,0)}
    +\operatorname{Broadcast}
    (W_{\mathrm{diff}}^{(b)}e_t+b_{\mathrm{diff}}^{(b)})$\;

    Apply temporal self-attention independently to each variable:
    $U_t^{(b,2)}
    \leftarrow
    \mathcal{T}_{\mathrm{time}}^{(b)}
    (U_t^{(b,1)})$\;

    Apply cross-variable self-attention independently at each
    timestep:
    $U_t^{(b,3)}
    \leftarrow
    \mathcal{T}_{\mathrm{var}}^{(b)}
    (U_t^{(b,2)})$\;

    Apply the pointwise hidden and side-information projections:
    $U_t^{(b,4)}
    \leftarrow
    \psi_{\mathrm{mid}}^{(b)}(U_t^{(b,3)})
    +\psi_{\mathrm{cond}}^{(b)}(S)$\;

    Split $U_t^{(b,4)}$ into
    $U_{t,\mathrm{gate}}^{(b,4)}$ and
    $U_{t,\mathrm{filter}}^{(b,4)}$\;

    Apply the gated activation:
    $U_t^{(b,5)}
    \leftarrow
    \sigma(U_{t,\mathrm{gate}}^{(b,4)})
    \odot
    \tanh(U_{t,\mathrm{filter}}^{(b,4)})$\;

    Apply the pointwise output projection:
    $U_t^{(b,6)}
    \leftarrow
    \psi_{\mathrm{out}}^{(b)}(U_t^{(b,5)})$\;

    Split $U_t^{(b,6)}$ into the residual update
    $U_{t,\mathrm{res}}^{(b)}$ and skip representation
    $U_{t,\mathrm{skip}}^{(b)}$\;

    Update the residual path:
    $U_t^{(b)}
    \leftarrow
    (U_t^{(b,0)}+U_{t,\mathrm{res}}^{(b)})/\sqrt{2}$\;

    Store $U_{t,\mathrm{skip}}^{(b)}$\;
}

Aggregate and normalize the skip representations:
$U_{t,\mathrm{skip}}
\leftarrow
\sum_{b=1}^{N_{\mathrm{blk}}}
U_{t,\mathrm{skip}}^{(b)}
/
\sqrt{N_{\mathrm{blk}}}$\;

Apply the first pointwise output projection:
$U_t^{\mathrm{out}}
\leftarrow
\operatorname{ReLU}
(\psi_{\mathrm{out},1}(U_{t,\mathrm{skip}}))$\;

Apply the final pointwise output projection:
$\widehat{\epsilon}_{\theta}
\leftarrow
\psi_{\mathrm{out},2}(U_t^{\mathrm{out}})$\;

\Return{$\widehat{\epsilon}_{\theta}$}\;

\end{algorithm}

\FloatBarrier

\section{Dataset-Specific \model{} Hyperparameters}
\label{app:rddmpi_hyperparameters}

Table~\ref{tab:rddmpi_hyperparameters} reports the selected
dataset-specific optimization and architectural hyperparameters
used to train \model{}. These include the batch size, learning
rate, number of residual denoising blocks, number of hidden channels, and
number of attention heads. The deterministic T1 backbone is
pretrained separately for each dataset and remains frozen during
residual diffusion training.

\begin{table}[h]
\centering
\setlength{\tabcolsep}{13.5pt}
\renewcommand{\arraystretch}{1.12}
\caption{Dataset-specific hyperparameters for \model{}.
$N_{\mathrm{blk}}$ denotes the number of residual denoising
blocks, and $d_h$ denotes the number of hidden channels.}
\label{tab:rddmpi_hyperparameters}
\begin{tabular}{lccccc}
\toprule
Dataset &
Batch Size &
Learning rate &
$N_{\mathrm{blk}}$ &
$d_h$ &
Number of heads \\
\midrule
ETTh1    & 16 & $10^{-3}$ & 4  & 128 & 8 \\
ETTh2    & 16 & $10^{-3}$ & 4 & 128  & 8 \\
Exchange & 16 & $10^{-3}$ & 4 & 128  & 8 \\
Illness  & 64 & $10^{-3}$ & 4 & 64  & 8 \\
Weather  & 16 & $10^{-4}$ & 4 & 64  & 8 \\
\bottomrule
\end{tabular}
\end{table}

\FloatBarrier

\section{Full Results}
\label{appendix:resultsl}

This section reports the complete quantitative results underlying the aggregated comparisons presented in the main paper. For point missingness, results are shown separately for missing ratios of 0.2, 0.4, 0.6, and 0.8, rather than averaging across these ratios as in the main tables. We also report the  results under block missingness.

All experiments are conducted using five random seeds: 2, 102, 202, 302, and 402. For each seed, the models are trained independently and evaluated on the corresponding artificially masked test data. MAE, MSE, and CRPS are first computed for each seed, after which we report their mean and standard deviation across the five runs. The mean tables summarize average performance, while the standard deviation tables quantify sensitivity to model initialization, training stochasticity, and the generated missingness patterns. We provide both the disaggregated results and their variability for transparency, reproducibility, and a more complete assessment of model robustness.

\begin{table}[h]
\centering
\caption{Full results under point missing ratios
(0.2, 0.4, 0.6, 0.8) across datasets. Best results are marked in
\textcolor{red}{\textbf{bold}}, and second-best results are marked
in \textcolor{blue}{\underline{underlined}}.}
\label{tab:point_missing_dataset_full}

\adjustbox{width=\textwidth}{%
\setlength{\tabcolsep}{2.5pt}

\begin{tabular}{l|l|*{10}{cc|}cc}
\toprule
\multicolumn{2}{c|}{Models}
& \multicolumn{2}{c|}{\textbf{\model{} (Ours)}}
& \multicolumn{2}{c|}{DLinear}
& \multicolumn{2}{c|}{ModernTCN}
& \multicolumn{2}{c|}{iTransformer}
& \multicolumn{2}{c|}{SAITS}
& \multicolumn{2}{c|}{ImputeFormer}
& \multicolumn{2}{c|}{TimesNet}
& \multicolumn{2}{c|}{T1}
& \multicolumn{2}{c|}{GP-VAE}
& \multicolumn{2}{c|}{CSDI}
& \multicolumn{2}{c}{FGTI} \\

\multicolumn{2}{c|}{Metric}
& MSE & MAE
& MSE & MAE
& MSE & MAE
& MSE & MAE
& MSE & MAE
& MSE & MAE
& MSE & MAE
& MSE & MAE
& MSE & MAE
& MSE & MAE
& MSE & MAE \\
\midrule

\multirow{5}{*}{\rotatebox{90}{ETTh1}}
& 0.2
& \textcolor{red}{\textbf{0.0229}}
& \textcolor{red}{\textbf{0.0956}}
& 0.1158 & 0.2320
& 0.0533 & 0.1594
& 0.0900 & 0.2033
& 0.0335 & 0.1180
& 0.0797 & 0.1697
& 0.0916 & 0.2075
& 0.0296 & 0.1113
& 0.3531 & 0.4602
& 0.0266 & 0.1016
& \textcolor{blue}{\underline{0.0253}}
& \textcolor{blue}{\underline{0.1009}} \\

& 0.4
& \textcolor{red}{\textbf{0.0337}}
& \textcolor{red}{\textbf{0.1123}}
& 0.1166 & 0.2232
& 0.0584 & 0.1632
& 0.1054 & 0.2172
& 0.0568 & 0.1482
& 0.1547 & 0.2232
& 0.0998 & 0.2123
& 0.0398 & 0.1277
& 0.5075 & 0.5365
& 0.0408 & 0.1225
& \textcolor{blue}{\underline{0.0391}}
& \textcolor{blue}{\underline{0.1220}} \\

& 0.6
& \textcolor{red}{\textbf{0.0480}}
& \textcolor{red}{\textbf{0.1362}}
& 0.2179 & 0.2927
& 0.0943 & 0.2020
& 0.1526 & 0.2533
& 0.1162 & 0.2097
& 0.2932 & 0.3138
& 0.1401 & 0.2434
& 0.0658 & 0.1632
& 0.6978 & 0.6235
& \textcolor{blue}{\underline{0.0610}}
& \textcolor{blue}{\underline{0.1512}}
& 0.0634 & 0.1537 \\

& 0.8
& \textcolor{red}{\textbf{0.1145}}
& \textcolor{red}{\textbf{0.2020}}
& 0.4234 & 0.4143
& 0.2520 & 0.3198
& 0.3105 & 0.3537
& 0.3337 & 0.3691
& 0.6149 & 0.5109
& 0.3220 & 0.3637
& 0.1878 & 0.2694
& 0.9399 & 0.7237
& \textcolor{blue}{\underline{0.1281}}
& \textcolor{blue}{\underline{0.2128}}
& 0.1594 & 0.2315 \\

\cmidrule(lr){2-24}
& Avg
& \textcolor{red}{\textbf{0.0548}}
& \textcolor{red}{\textbf{0.1365}}
& 0.2184 & 0.2906
& 0.1145 & 0.2111
& 0.1646 & 0.2569
& 0.1351 & 0.2113
& 0.2856 & 0.3044
& 0.1634 & 0.2567
& 0.0807 & 0.1679
& 0.6246 & 0.5860
& \textcolor{blue}{\underline{0.0641}}
& \textcolor{blue}{\underline{0.1470}}
& 0.0718 & 0.1520 \\
\midrule

\multirow{5}{*}{\rotatebox{90}{ETTh2}}
& 0.2
& \textcolor{red}{\textbf{0.0228}}
& \textcolor{red}{\textbf{0.0771}}
& 0.0562 & 0.1578
& 0.0390 & 0.1269
& 0.0495 & 0.1463
& 0.1442 & 0.2685
& 0.1490 & 0.2279
& 0.0524 & 0.1542
& \textcolor{blue}{\underline{0.0255}}
& \textcolor{blue}{\underline{0.0954}}
& 0.6106 & 0.5741
& 0.0415 & 0.1008
& 0.0402 & 0.1149 \\

& 0.4
& \textcolor{red}{\textbf{0.0305}}
& \textcolor{red}{\textbf{0.0924}}
& 0.0557 & 0.1560
& 0.0403 & 0.1286
& 0.0560 & 0.1553
& 0.1788 & 0.2950
& 0.2451 & 0.2845
& 0.0559 & 0.1577
& \textcolor{blue}{\underline{0.0310}}
& \textcolor{blue}{\underline{0.1057}}
& 0.8039 & 0.6763
& 0.0532 & 0.1206
& 0.0502 & 0.1313 \\

& 0.6
& \textcolor{blue}{\underline{0.0437}}
& \textcolor{red}{\textbf{0.1150}}
& 0.0799 & 0.1871
& 0.0540 & 0.1501
& 0.0693 & 0.1744
& 0.3431 & 0.3897
& 0.6173 & 0.4458
& 0.0729 & 0.1787
& \textcolor{red}{\textbf{0.0437}}
& \textcolor{blue}{\underline{0.1283}}
& 1.4455 & 0.9409
& 0.0712 & 0.1475
& 0.0726 & 0.1594 \\

& 0.8
& \textcolor{blue}{\underline{0.0797}}
& \textcolor{red}{\textbf{0.1649}}
& 0.1287 & 0.2396
& 0.0973 & 0.2043
& 0.1099 & 0.2219
& 1.0818 & 0.6869
& 1.6554 & 0.8247
& 0.1182 & 0.2291
& \textcolor{red}{\textbf{0.0764}}
& \textcolor{blue}{\underline{0.1764}}
& 2.7956 & 1.2937
& 0.1280 & 0.2089
& 0.2299 & 0.2728 \\

\cmidrule(lr){2-24}
& Avg
& \textcolor{red}{\textbf{0.0442}}
& \textcolor{red}{\textbf{0.1123}}
& 0.0801 & 0.1851
& 0.0576 & 0.1524
& 0.0712 & 0.1745
& 0.4370 & 0.4100
& 0.6667 & 0.4457
& 0.0749 & 0.1799
& \textcolor{blue}{\underline{0.0442}}
& \textcolor{blue}{\underline{0.1264}}
& 1.4139 & 0.8712
& 0.0735 & 0.1444
& 0.0982 & 0.1696 \\
\midrule

\multirow{5}{*}{\rotatebox{90}{Exchange}}
& 0.2
& 0.0022
& \textcolor{red}{\textbf{0.0147}}
& 0.0029 & 0.0364
& 0.0053 & 0.0512
& \textcolor{blue}{\underline{0.0016}} & 0.0262
& 0.1106 & 0.2863
& 0.0115 & 0.0401
& 0.0017 & 0.0258
& \textcolor{red}{\textbf{0.0007}}
& \textcolor{blue}{\underline{0.0151}}
& 0.5095 & 0.6226
& 0.0315 & 0.0889
& 0.0029 & 0.0250 \\

& 0.4
& \textcolor{blue}{\underline{0.0015}}
& \textcolor{red}{\textbf{0.0171}}
& 0.0025 & 0.0317
& 0.0075 & 0.0602
& 0.0021 & 0.0278
& 0.1480 & 0.3225
& 0.0134 & 0.0493
& 0.0021 & 0.0266
& \textcolor{red}{\textbf{0.0012}}
& \textcolor{blue}{\underline{0.0174}}
& 0.5551 & 0.6523
& 0.0382 & 0.1008
& 0.0021 & 0.0279 \\

& 0.6
& \textcolor{red}{\textbf{0.0019}}
& \textcolor{red}{\textbf{0.0206}}
& 0.0053 & 0.0453
& 0.0109 & 0.0717
& 0.0032 & 0.0328
& 0.2328 & 0.3882
& 0.0451 & 0.0951
& 0.0033 & 0.0321
& \textcolor{blue}{\underline{0.0020}}
& \textcolor{blue}{\underline{0.0217}}
& 0.6376 & 0.6981
& 0.0472 & 0.1147
& 0.0034 & 0.0355 \\

& 0.8
& \textcolor{red}{\textbf{0.0030}}
& \textcolor{red}{\textbf{0.0288}}
& 0.0105 & 0.0680
& 0.0144 & 0.0823
& 0.0064 & 0.0488
& 0.4330 & 0.5289
& 0.2070 & 0.2549
& 0.0067 & 0.0496
& \textcolor{blue}{\underline{0.0038}}
& \textcolor{blue}{\underline{0.0332}}
& 0.8640 & 0.8046
& 0.0772 & 0.1496
& 0.0142 & 0.0753 \\

\cmidrule(lr){2-24}
& Avg
& \textcolor{blue}{\underline{0.0022}}
& \textcolor{red}{\textbf{0.0203}}
& 0.0053 & 0.0453
& 0.0095 & 0.0664
& 0.0033 & 0.0339
& 0.2311 & 0.3815
& 0.0693 & 0.1098
& 0.0034 & 0.0335
& \textcolor{red}{\textbf{0.0019}}
& \textcolor{blue}{\underline{0.0218}}
& 0.6415 & 0.6944
& 0.0485 & 0.1135
& 0.0056 & 0.0409 \\
\midrule

\multirow{5}{*}{\rotatebox{90}{Illness}}
& 0.2
& \textcolor{red}{\textbf{0.0057}}
& \textcolor{red}{\textbf{0.0449}}
& 0.1392 & 0.2691
& 0.0236 & 0.1078
& 0.0722 & 0.1724
& 0.1344 & 0.2193
& 0.2115 & 0.2625
& 0.0434 & 0.1468
& \textcolor{blue}{\underline{0.0079}}
& \textcolor{blue}{\underline{0.0561}}
& 0.2975 & 0.3760
& 0.0394 & 0.1055
& 0.0288 & 0.0918 \\

& 0.4
& \textcolor{red}{\textbf{0.0081}}
& \textcolor{red}{\textbf{0.0518}}
& 0.0790 & 0.1842
& 0.0261 & 0.1079
& 0.0860 & 0.1985
& 0.1772 & 0.2448
& 0.2690 & 0.3014
& 0.0394 & 0.1390
& \textcolor{blue}{\underline{0.0095}}
& \textcolor{blue}{\underline{0.0597}}
& 0.5426 & 0.4767
& 0.0506 & 0.1160
& 0.0362 & 0.1045 \\

& 0.6
& \textcolor{red}{\textbf{0.0132}}
& \textcolor{red}{\textbf{0.0664}}
& 0.1781 & 0.2829
& 0.0451 & 0.1447
& 0.0965 & 0.2093
& 0.2600 & 0.3040
& 0.3376 & 0.3512
& 0.0670 & 0.1818
& \textcolor{blue}{\underline{0.0156}}
& \textcolor{blue}{\underline{0.0759}}
& 0.7072 & 0.5588
& 0.0772 & 0.1525
& 0.0660 & 0.1452 \\

& 0.8
& \textcolor{red}{\textbf{0.0673}}
& \textcolor{red}{\textbf{0.1225}}
& 0.4736 & 0.4823
& 0.2394 & 0.3339
& 0.2298 & 0.3001
& 0.5777 & 0.4493
& 0.5394 & 0.4542
& 0.2377 & 0.3541
& \textcolor{blue}{\underline{0.0676}}
& \textcolor{blue}{\underline{0.1479}}
& 0.9947 & 0.6593
& 0.1550 & 0.2178
& 0.2128 & 0.2441 \\

\cmidrule(lr){2-24}
& Avg
& \textcolor{red}{\textbf{0.0236}}
& \textcolor{red}{\textbf{0.0714}}
& 0.2175 & 0.3046
& 0.0835 & 0.1736
& 0.1211 & 0.2201
& 0.2873 & 0.3044
& 0.3394 & 0.3423
& 0.0969 & 0.2054
& \textcolor{blue}{\underline{0.0252}}
& \textcolor{blue}{\underline{0.0849}}
& 0.6355 & 0.5177
& 0.0806 & 0.1479
& 0.0860 & 0.1464 \\
\midrule

\multirow{5}{*}{\rotatebox{90}{Weather}}
& 0.2
& \textcolor{red}{\textbf{0.0246}}
& \textcolor{red}{\textbf{0.0246}}
& 0.0354 & 0.0745
& 0.0305 & 0.0632
& 0.0911 & 0.1436
& 0.0252 & 0.0308
& 0.0324 & 0.0355
& 0.0351 & 0.0729
& \textcolor{blue}{\underline{0.0249}} & 0.0375
& 0.0806 & 0.1582
& 0.0269 & 0.0253
& 0.0255
& \textcolor{blue}{\underline{0.0252}} \\

& 0.4
& \textcolor{blue}{\underline{0.0271}}
& \textcolor{red}{\textbf{0.0271}}
& 0.0316 & 0.0585
& 0.0302 & 0.0563
& 0.0873 & 0.1431
& 0.0272 & 0.0359
& 0.0329 & 0.0392
& 0.0348 & 0.0685
& \textcolor{red}{\textbf{0.0268}} & 0.0407
& 0.1201 & 0.2048
& 0.0289
& \textcolor{blue}{\underline{0.0277}}
& 0.0285 & 0.0278 \\

& 0.6
& \textcolor{red}{\textbf{0.0313}}
& \textcolor{red}{\textbf{0.0317}}
& 0.0449 & 0.0863
& 0.0401 & 0.0773
& 0.0900 & 0.1443
& 0.0387 & 0.0557
& 0.0409 & 0.0514
& 0.0444 & 0.0880
& 0.0330 & 0.0529
& 0.1859 & 0.2753
& \textcolor{blue}{\underline{0.0319}}
& \textcolor{blue}{\underline{0.0317}}
& 0.0346 & 0.0323 \\

& 0.8
& \textcolor{blue}{\underline{0.0409}}
& \textcolor{blue}{\underline{0.0425}}
& 0.0769 & 0.1334
& 0.0692 & 0.1228
& 0.0998 & 0.1497
& 0.0902 & 0.1363
& 0.0839 & 0.1114
& 0.0743 & 0.1317
& 0.0606 & 0.1016
& 0.3206 & 0.3927
& \textcolor{red}{\textbf{0.0408}}
& \textcolor{red}{\textbf{0.0415}}
& 0.0457 & 0.0428 \\

\cmidrule(lr){2-24}
& Avg
& \textcolor{red}{\textbf{0.0310}}
& \textcolor{red}{\textbf{0.0315}}
& 0.0472 & 0.0882
& 0.0425 & 0.0799
& 0.0920 & 0.1452
& 0.0453 & 0.0647
& 0.0475 & 0.0594
& 0.0471 & 0.0903
& 0.0363 & 0.0582
& 0.1768 & 0.2578
& \textcolor{blue}{\underline{0.0321}}
& \textcolor{blue}{\underline{0.0316}}
& 0.0335 & 0.0320 \\

\bottomrule
\end{tabular}%
}
\end{table}

\begin{table*}[t]
\centering
\caption{The standard deviation of Table~\ref{tab:point_missing_dataset_full}.}
\label{tab:std_point_missing_dataset_std}

\adjustbox{width=\textwidth}{%
\setlength{\tabcolsep}{2.5pt}

\begin{tabular}{l|l|*{10}{cc|}cc}
\toprule
\multicolumn{2}{c|}{Models}
& \multicolumn{2}{c|}{\textbf{\model{} (Ours)}}
& \multicolumn{2}{c|}{DLinear}
& \multicolumn{2}{c|}{ModernTCN}
& \multicolumn{2}{c|}{iTransformer}
& \multicolumn{2}{c|}{SAITS}
& \multicolumn{2}{c|}{ImputeFormer}
& \multicolumn{2}{c|}{TimesNet}
& \multicolumn{2}{c|}{T1}
& \multicolumn{2}{c|}{GP-VAE}
& \multicolumn{2}{c|}{CSDI}
& \multicolumn{2}{c}{FGTI} \\

\multicolumn{2}{c|}{Metric}
& MSE & MAE
& MSE & MAE
& MSE & MAE
& MSE & MAE
& MSE & MAE
& MSE & MAE
& MSE & MAE
& MSE & MAE
& MSE & MAE
& MSE & MAE
& MSE & MAE \\
\midrule

\multirow{5}{*}{\rotatebox{90}{ETTh1}}
& 0.2
& 0.0025 & 0.0008
& 0.0071 & 0.0047
& 0.0026 & 0.0026
& 0.0038 & 0.0023
& 0.0046 & 0.0068
& 0.0087 & 0.0080
& 0.0064 & 0.0081
& 0.0014 & 0.0017
& 0.0086 & 0.0057
& 0.0022 & 0.0031
& 0.0017 & 0.0017 \\

& 0.4
& 0.0028 & 0.0015
& 0.0068 & 0.0039
& 0.0035 & 0.0028
& 0.0033 & 0.0032
& 0.0137 & 0.0125
& 0.0184 & 0.0116
& 0.0028 & 0.0036
& 0.0024 & 0.0017
& 0.0130 & 0.0065
& 0.0058 & 0.0044
& 0.0045 & 0.0035 \\

& 0.6
& 0.0023 & 0.0015
& 0.0167 & 0.0065
& 0.0070 & 0.0045
& 0.0112 & 0.0046
& 0.0318 & 0.0241
& 0.0421 & 0.0281
& 0.0133 & 0.0051
& 0.0036 & 0.0023
& 0.0119 & 0.0057
& 0.0034 & 0.0040
& 0.0058 & 0.0048 \\

& 0.8
& 0.0081 & 0.0051
& 0.0141 & 0.0071
& 0.0156 & 0.0089
& 0.0195 & 0.0064
& 0.0501 & 0.0367
& 0.0613 & 0.0457
& 0.0206 & 0.0120
& 0.0074 & 0.0066
& 0.0136 & 0.0093
& 0.0218 & 0.0106
& 0.0171 & 0.0096 \\

\cmidrule(lr){2-24}
& Avg
& 0.0039 & 0.0022
& 0.0111 & 0.0056
& 0.0072 & 0.0047
& 0.0095 & 0.0041
& 0.0251 & 0.0200
& 0.0326 & 0.0233
& 0.0108 & 0.0072
& 0.0037 & 0.0031
& 0.0118 & 0.0068
& 0.0083 & 0.0055
& 0.0073 & 0.0049 \\
\midrule

\multirow{5}{*}{\rotatebox{90}{ETTh2}}
& 0.2
& 0.0012 & 0.0034
& 0.0030 & 0.0024
& 0.0022 & 0.0027
& 0.0027 & 0.0023
& 0.0071 & 0.0078
& 0.0126 & 0.0056
& 0.0029 & 0.0026
& 0.0012 & 0.0015
& 0.0730 & 0.0339
& 0.0052 & 0.0063
& 0.0037 & 0.0054 \\

& 0.4
& 0.0022 & 0.0030
& 0.0014 & 0.0013
& 0.0003 & 0.0006
& 0.0003 & 0.0005
& 0.0163 & 0.0152
& 0.0429 & 0.0221
& 0.0011 & 0.0011
& 0.0012 & 0.0011
& 0.0784 & 0.0389
& 0.0085 & 0.0084
& 0.0016 & 0.0029 \\

& 0.6
& 0.0044 & 0.0049
& 0.0010 & 0.0016
& 0.0008 & 0.0018
& 0.0016 & 0.0019
& 0.0937 & 0.0546
& 0.1761 & 0.0650
& 0.0014 & 0.0028
& 0.0015 & 0.0018
& 0.1417 & 0.0374
& 0.0112 & 0.0093
& 0.0035 & 0.0027 \\

& 0.8
& 0.0081 & 0.0079
& 0.0035 & 0.0024
& 0.0041 & 0.0034
& 0.0035 & 0.0028
& 0.2958 & 0.1241
& 0.3109 & 0.0934
& 0.0021 & 0.0020
& 0.0041 & 0.0037
& 0.0974 & 0.0177
& 0.0285 & 0.0192
& 0.0792 & 0.0425 \\

\cmidrule(lr){2-24}
& Avg
& 0.0040 & 0.0048
& 0.0022 & 0.0019
& 0.0019 & 0.0021
& 0.0020 & 0.0019
& 0.1032 & 0.0504
& 0.1356 & 0.0465
& 0.0019 & 0.0021
& 0.0020 & 0.0020
& 0.0976 & 0.0320
& 0.0133 & 0.0108
& 0.0220 & 0.0134 \\
\midrule

\multirow{5}{*}{\rotatebox{90}{Exchange}}
& 0.2
& 0.0019 & 0.0003
& 0.0002 & 0.0005
& 0.0004 & 0.0010
& 0.0002 & 0.0002
& 0.0154 & 0.0221
& 0.0084 & 0.0035
& 0.0002 & 0.0009
& 0.0002 & 0.0004
& 0.0400 & 0.0259
& 0.0383 & 0.0789
& 0.0019 & 0.0009 \\

& 0.4
& 0.0008 & 0.0005
& 0.0008 & 0.0003
& 0.0008 & 0.0005
& 0.0009 & 0.0003
& 0.0056 & 0.0082
& 0.0054 & 0.0117
& 0.0007 & 0.0007
& 0.0008 & 0.0003
& 0.0300 & 0.0197
& 0.0548 & 0.0970
& 0.0011 & 0.0030 \\

& 0.6
& 0.0005 & 0.0003
& 0.0006 & 0.0009
& 0.0006 & 0.0005
& 0.0006 & 0.0006
& 0.0106 & 0.0130
& 0.0417 & 0.0532
& 0.0004 & 0.0007
& 0.0006 & 0.0004
& 0.0373 & 0.0208
& 0.0698 & 0.1167
& 0.0006 & 0.0052 \\

& 0.8
& 0.0001 & 0.0002
& 0.0004 & 0.0007
& 0.0003 & 0.0005
& 0.0004 & 0.0011
& 0.0229 & 0.0210
& 0.2059 & 0.1700
& 0.0005 & 0.0010
& 0.0006 & 0.0012
& 0.0497 & 0.0251
& 0.1148 & 0.1518
& 0.0059 & 0.0203 \\

\cmidrule(lr){2-24}
& Avg
& 0.0008 & 0.0003
& 0.0005 & 0.0006
& 0.0005 & 0.0007
& 0.0005 & 0.0006
& 0.0136 & 0.0161
& 0.0653 & 0.0596
& 0.0005 & 0.0008
& 0.0005 & 0.0006
& 0.0392 & 0.0229
& 0.0694 & 0.1111
& 0.0024 & 0.0073 \\
\midrule

\multirow{5}{*}{\rotatebox{90}{Illness}}
& 0.2 & 0.0007 & 0.0029 & 0.0159 & 0.0107 & 0.0043 & 0.0120 & 0.0200 & 0.0249 & 0.0327 & 0.0314 & 0.0447 & 0.0276 & 0.0052 & 0.0132 & 0.0018 & 0.0061 & 0.0642 & 0.0333 & 0.0075 & 0.0088 & 0.0045 & 0.0104 \\

& 0.4 & 0.0014 & 0.0028 & 0.0098 & 0.0091 & 0.0031 & 0.0047 & 0.0077 & 0.0049 & 0.0385 & 0.0121 & 0.0768 & 0.0367 & 0.0050 & 0.0073 & 0.0013 & 0.0016 & 0.0805 & 0.0268 & 0.0090 & 0.0083 & 0.0088 & 0.0117 \\

& 0.6 & 0.0020 & 0.0036 & 0.0514 & 0.0329 & 0.0089 & 0.0134 & 0.0228 & 0.0181 & 0.0445 & 0.0182 & 0.0384 & 0.0161 & 0.0135 & 0.0200 & 0.0022 & 0.0028 & 0.0734 & 0.0267 & 0.0153 & 0.0127 & 0.0145 & 0.0169 \\

& 0.8 & 0.0389 & 0.0138 & 0.0858 & 0.0268 & 0.0644 & 0.0334 & 0.0728 & 0.0316 & 0.1379 & 0.0425 & 0.1117 & 0.0419 & 0.0426 & 0.0230 & 0.0387 & 0.0271 & 0.0599 & 0.0184 & 0.0290 & 0.0131 & 0.0891 & 0.0312 \\

\cmidrule(lr){2-24}
& Avg & 0.0108 & 0.0058 & 0.0407 & 0.0199 & 0.0202 & 0.0159 & 0.0308 & 0.0199 & 0.0634 & 0.0260 & 0.0679 & 0.0306 & 0.0166 & 0.0159 & 0.0110 & 0.0094 & 0.0695 & 0.0263 & 0.0152 & 0.0107 & 0.0292 & 0.0175 \\
\midrule

\multirow{5}{*}{\rotatebox{90}{Weather}}
& 0.2 & 0.0018 & 0.0003 & 0.0026 & 0.0013 & 0.0010 & 0.0055 & 0.0040 & 0.0013 & 0.0022 & 0.0012 & 0.0039 & 0.0034 & 0.0014 & 0.0037 & 0.0017 & 0.0025 & 0.0079 & 0.0096 & 0.0023 & 0.0006 & 0.0014 & 0.0022 \\

& 0.4 & 0.0020 & 0.0006 & 0.0007 & 0.0004 & 0.0006 & 0.0012 & 0.0018 & 0.0005 & 0.0016 & 0.0018 & 0.0039 & 0.0036 & 0.0011 & 0.0026 & 0.0022 & 0.0032 & 0.0081 & 0.0087 & 0.0022 & 0.0010 & 0.0027 & 0.0028 \\

& 0.6 & 0.0020 & 0.0014 & 0.0015 & 0.0003 & 0.0025 & 0.0040 & 0.0019 & 0.0002 & 0.0022 & 0.0033 & 0.0004 & 0.0047 & 0.0005 & 0.0028 & 0.0027 & 0.0058 & 0.0101 & 0.0109 & 0.0024 & 0.0012 & 0.0027 & 0.0032 \\

& 0.8 & 0.0014 & 0.0022 & 0.0004 & 0.0005 & 0.0020 & 0.0046 & 0.0042 & 0.0013 & 0.0079 & 0.0116 & 0.0161 & 0.0219 & 0.0011 & 0.0015 & 0.0043 & 0.0106 & 0.0258 & 0.0244 & 0.0007 & 0.0014 & 0.0056 & 0.0048 \\

\cmidrule(lr){2-24}
& Avg & 0.0018 & 0.0011 & 0.0013 & 0.0006 & 0.0015 & 0.0038 & 0.0030 & 0.0008 & 0.0035 & 0.0045 & 0.0061 & 0.0084 & 0.0010 & 0.0026 & 0.0027 & 0.0055 & 0.0130 & 0.0134 & 0.0019 & 0.0010 & 0.0031 & 0.0033 \\

\bottomrule
\end{tabular}%
}
\end{table*}

\begin{table*}[t]
\centering
\caption{Full results under the block missing scenario across
datasets. Best results are marked in
\textcolor{red}{\textbf{bold}}, and second-best results are marked
in \textcolor{blue}{\underline{underlined}}.}
\label{tab:block_missing_dataset_full}

\adjustbox{width=\textwidth}{%
\setlength{\tabcolsep}{2.5pt}

\begin{tabular}{l|*{10}{cc|}cc}
\toprule
\multirow{2}{*}{Dataset}
& \multicolumn{2}{c|}{\textbf{\model{} (Ours)}}
& \multicolumn{2}{c|}{DLinear}
& \multicolumn{2}{c|}{ModernTCN}
& \multicolumn{2}{c|}{iTransformer}
& \multicolumn{2}{c|}{SAITS}
& \multicolumn{2}{c|}{ImputeFormer}
& \multicolumn{2}{c|}{TimesNet}
& \multicolumn{2}{c|}{T1}
& \multicolumn{2}{c|}{GP-VAE}
& \multicolumn{2}{c|}{CSDI}
& \multicolumn{2}{c}{FGTI} \\

& MSE & MAE
& MSE & MAE
& MSE & MAE
& MSE & MAE
& MSE & MAE
& MSE & MAE
& MSE & MAE
& MSE & MAE
& MSE & MAE
& MSE & MAE
& MSE & MAE \\
\midrule

ETTh1
& \textcolor{red}{\textbf{0.0168}}
& \textcolor{red}{\textbf{0.0851}}
& 0.1728 & 0.2844
& 0.0592 & 0.1735
& 0.1016 & 0.2097
& 0.0257 & 0.1075
& 0.0594 & 0.1543
& 0.0920 & 0.2117
& 0.0258 & 0.1077
& 0.2842 & 0.4231
& 0.0179 & 0.0896
& \textcolor{blue}{\underline{0.0170}}
& \textcolor{blue}{\underline{0.0877}} \\

ETTh2
& \textcolor{red}{\textbf{0.0203}}
& \textcolor{red}{\textbf{0.0744}}
& 0.0770 & 0.1890
& 0.0477 & 0.1414
& 0.0546 & 0.1552
& 0.1471 & 0.2711
& 0.2654 & 0.2672
& 0.0533 & 0.1592
& \textcolor{blue}{\underline{0.0292}}
& \textcolor{blue}{\underline{0.1014}}
& 0.6001 & 0.5663
& 0.0561 & 0.1070
& 0.0948 & 0.1321 \\

Exchange
& 0.0043
& \textcolor{red}{\textbf{0.0181}}
& 0.0063 & 0.0557
& 0.0054 & 0.0498
& \textcolor{red}{\textbf{0.0034}} & 0.0336
& 0.1864 & 0.3330
& 0.1217 & 0.1233
& \textcolor{blue}{\underline{0.0036}} & 0.0362
& 0.0115
& \textcolor{blue}{\underline{0.0312}}
& 0.5100 & 0.6290
& 0.2237 & 0.1458
& 0.0177 & 0.0457 \\

Illness & \textcolor{red}{\textbf{0.0390}} & \textcolor{red}{\textbf{0.1282}} & 0.2603 & 0.3691 & 0.1521 & 0.2647 & 0.3345 & 0.3598 & 0.1475 & 0.2373 & 0.2697 & 0.3100 & 0.1368 & 0.2535 & 0.0874 & 0.1792 & 0.3270 & 0.4122 & 0.0977 & 0.1895 & \textcolor{blue}{\underline{0.0516}} & \textcolor{blue}{\underline{0.1283}} \\

Weather & \textcolor{red}{\textbf{0.0233}} & \textcolor{red}{\textbf{0.0254}} & 0.0495 & 0.1053 & 0.0371 & 0.0827 & 0.1003 & 0.1480 & 0.0251 & 0.0328 & 0.0401 & 0.0461 & 0.0390 & 0.0855 & 0.0248 & 0.0400 & 0.0627 & 0.1349 & 0.0234 & \textcolor{blue}{\underline{0.0254}} & \textcolor{blue}{\underline{0.0236}} & 0.0256 \\

\bottomrule
\end{tabular}%
}
\end{table*}

\begin{table*}[t]
\centering
\caption{Standard deviations corresponding to the block missing
results reported in
Table~\ref{tab:block_missing_dataset_full}.}
\label{tab:std_block_missing_dataset_full}

\adjustbox{width=\textwidth}{%
\setlength{\tabcolsep}{2.5pt}

\begin{tabular}{l|*{10}{cc|}cc}
\toprule
\multirow{2}{*}{Dataset}
& \multicolumn{2}{c|}{\textbf{\model{} (Ours)}}
& \multicolumn{2}{c|}{DLinear}
& \multicolumn{2}{c|}{ModernTCN}
& \multicolumn{2}{c|}{iTransformer}
& \multicolumn{2}{c|}{SAITS}
& \multicolumn{2}{c|}{ImputeFormer}
& \multicolumn{2}{c|}{TimesNet}
& \multicolumn{2}{c|}{T1}
& \multicolumn{2}{c|}{GP-VAE}
& \multicolumn{2}{c|}{CSDI}
& \multicolumn{2}{c}{FGTI} \\

& MSE & MAE
& MSE & MAE
& MSE & MAE
& MSE & MAE
& MSE & MAE
& MSE & MAE
& MSE & MAE
& MSE & MAE
& MSE & MAE
& MSE & MAE
& MSE & MAE \\
\midrule

ETTh1
& 0.0020 & 0.0025
& 0.0050 & 0.0024
& 0.0066 & 0.0060
& 0.0345 & 0.0129
& 0.0061 & 0.0100
& 0.0155 & 0.0130
& 0.0085 & 0.0077
& 0.0051 & 0.0049
& 0.0114 & 0.0067
& 0.0022 & 0.0046
& 0.0020 & 0.0041 \\

ETTh2
& 0.0024 & 0.0043
& 0.0099 & 0.0143
& 0.0117 & 0.0163
& 0.0102 & 0.0132
& 0.0168 & 0.0148
& 0.1646 & 0.0641
& 0.0070 & 0.0109
& 0.0078 & 0.0129
& 0.0693 & 0.0356
& 0.0324 & 0.0202
& 0.1091 & 0.0341 \\

Exchange
& 0.0074 & 0.0091
& 0.0012 & 0.0027
& 0.0020 & 0.0060
& 0.0018 & 0.0049
& 0.0852 & 0.0602
& 0.1218 & 0.0673
& 0.0014 & 0.0042
& 0.0192 & 0.0204
& 0.0608 & 0.0364
& 0.3921 & 0.1583
& 0.0322 & 0.0363 \\

Illness & 0.0384 & 0.0675 & 0.1112 & 0.0515 & 0.0985 & 0.0867 & 0.2518 & 0.1288 & 0.0663 & 0.0893 & 0.1174 & 0.0934 & 0.1194 & 0.1058 & 0.0885 & 0.1076 & 0.1543 & 0.1103 & 0.0984 & 0.1226 & 0.0719 & 0.1089 \\

Weather & 0.0041 & 0.0024 & 0.0062 & 0.0041 & 0.0050 & 0.0040 & 0.0140 & 0.0059 & 0.0061 & 0.0040 & 0.0023 & 0.0025 & 0.0047 & 0.0020 & 0.0052 & 0.0031 & 0.0084 & 0.0102 & 0.0051 & 0.0026 & 0.0032 & 0.0039 \\

\bottomrule
\end{tabular}%
}
\end{table*}

\begin{table*}[t]
\centering
\caption{Full CRPS comparison of \model{}, GP-VAE, CSDI, and
FGTI under point missing and block missing scenarios across datasets.
For point missingness, results are reported at missing ratios
0.2, 0.4, 0.6, and 0.8. Best results are marked in
\textcolor{red}{\textbf{bold}}, and second-best results are marked in
\textcolor{blue}{\underline{underlined}}. Lower is better.}
\label{tab:crps_point_block}
\vspace{-1ex}

\begin{minipage}[t]{0.66\textwidth}
\centering
\caption*{(a) Point missing}
\setlength{\tabcolsep}{4pt}
\renewcommand{\arraystretch}{0.95}
\begin{tabular}{l|l|cccc}
\toprule
\multicolumn{2}{c|}{Models}
& \textbf{\model{}}
& GP-VAE
& CSDI
& FGTI \\
\multicolumn{2}{c|}{Metric}
& CRPS & CRPS & CRPS & CRPS \\
\midrule

\multirow{5}{*}{\rotatebox{90}{ETTh1}}
& 0.2
& \textcolor{red}{\textbf{0.0924}}
& 0.5829
& 0.0970
& \textcolor{blue}{\underline{0.0966}} \\
& 0.4
& \textcolor{red}{\textbf{0.1080}}
& 0.6739
& 0.1165
& \textcolor{blue}{\underline{0.1161}} \\
& 0.6
& \textcolor{red}{\textbf{0.1310}}
& 0.7856
& \textcolor{blue}{\underline{0.1446}}
& 0.1469 \\
& 0.8
& \textcolor{red}{\textbf{0.1940}}
& 0.9080
& \textcolor{blue}{\underline{0.2044}}
& 0.2222 \\
\cmidrule(lr){2-6}
& Avg
& \textcolor{red}{\textbf{0.1313}}
& 0.7376
& \textcolor{blue}{\underline{0.1406}}
& 0.1454 \\
\midrule

\multirow{5}{*}{\rotatebox{90}{ETTh2}}
& 0.2
& \textcolor{red}{\textbf{0.0438}}
& 0.4202
& \textcolor{blue}{\underline{0.0570}}
& 0.0657 \\
& 0.4
& \textcolor{red}{\textbf{0.0528}}
& 0.4967
& \textcolor{blue}{\underline{0.0682}}
& 0.0756 \\
& 0.6
& \textcolor{red}{\textbf{0.0657}}
& 0.6921
& \textcolor{blue}{\underline{0.0848}}
& 0.0928 \\
& 0.8
& \textcolor{red}{\textbf{0.0941}}
& 0.9488
& \textcolor{blue}{\underline{0.1203}}
& 0.1576 \\
\cmidrule(lr){2-6}
& Avg
& \textcolor{red}{\textbf{0.0641}}
& 0.6395
& \textcolor{blue}{\underline{0.0826}}
& 0.0979 \\
\midrule

\multirow{5}{*}{\rotatebox{90}{Exchange}}
& 0.2
& \textcolor{red}{\textbf{0.0131}}
& 0.7039
& 0.0766
& \textcolor{blue}{\underline{0.0216}} \\
& 0.4
& \textcolor{red}{\textbf{0.0151}}
& 0.7360
& 0.0869
& \textcolor{blue}{\underline{0.0239}} \\
& 0.6
& \textcolor{red}{\textbf{0.0181}}
& 0.7887
& 0.0998
& \textcolor{blue}{\underline{0.0304}} \\
& 0.8
& \textcolor{red}{\textbf{0.0252}}
& 0.9106
& 0.1328
& \textcolor{blue}{\underline{0.0658}} \\
\cmidrule(lr){2-6}
& Avg
& \textcolor{red}{\textbf{0.0179}}
& 0.7848
& 0.0990
& \textcolor{blue}{\underline{0.0354}} \\
\midrule

\multirow{5}{*}{\rotatebox{90}{Illness}}
& 0.2
& \textcolor{red}{\textbf{0.0460}}
& 0.4830
& 0.1111
& \textcolor{blue}{\underline{0.0917}} \\
& 0.4
& \textcolor{red}{\textbf{0.0526}}
& 0.6118
& 0.1209
& \textcolor{blue}{\underline{0.1026}} \\
& 0.6
& \textcolor{red}{\textbf{0.0680}}
& 0.7244
& 0.1617
& \textcolor{blue}{\underline{0.1412}} \\
& 0.8
& \textcolor{red}{\textbf{0.1248}}
& 0.8479
& \textcolor{blue}{\underline{0.2296}}
& 0.2427 \\
\cmidrule(lr){2-6}
& Avg
& \textcolor{red}{\textbf{0.0728}}
& 0.6668
& 0.1558
& \textcolor{blue}{\underline{0.1445}} \\
\midrule

\multirow{5}{*}{\rotatebox{90}{Weather}}
& 0.2
& \textcolor{red}{\textbf{0.0327}}
& 0.2764
& \textcolor{blue}{\underline{0.0335}}
& 0.0365 \\
& 0.4
& \textcolor{red}{\textbf{0.0362}}
& 0.3575
& \textcolor{blue}{\underline{0.0368}}
& 0.0400 \\
& 0.6
& \textcolor{blue}{\underline{0.0423}}
& 0.4809
& \textcolor{red}{\textbf{0.0421}}
& 0.0468 \\
& 0.8
& \textcolor{blue}{\underline{0.0573}}
& 0.6855
& \textcolor{red}{\textbf{0.0555}}
& 0.0626 \\
\cmidrule(lr){2-6}
& Avg
& \textcolor{red}{\textbf{0.0420}}
& 0.4501
& \textcolor{blue}{\underline{0.0421}}
& 0.0465 \\
\bottomrule
\end{tabular}
\end{minipage}
\hfill
\begin{minipage}[t]{0.30\textwidth}
\centering
\caption*{(b) Block missing}
\setlength{\tabcolsep}{3pt}
\renewcommand{\arraystretch}{1.0}
\begin{tabular}{lcccc}
\toprule
Dataset & \textbf{\model{}} & GP-VAE & CSDI & FGTI \\
\midrule

ETTh1
& \textcolor{red}{\textbf{0.0811}}
& 0.5298
& 0.0867
& \textcolor{blue}{\underline{0.0832}} \\

ETTh2
& \textcolor{red}{\textbf{0.0412}}
& 0.4076
& \textcolor{blue}{\underline{0.0596}}
& 0.0751 \\

Exchange
& \textcolor{red}{\textbf{0.0153}}
& 0.6903
& 0.1238
& \textcolor{blue}{\underline{0.0391}} \\

Illness
& \textcolor{blue}{\underline{0.1238}}
& 0.4793
& 0.1873
& \textcolor{red}{\textbf{0.1173}} \\

Weather
& \textcolor{blue}{\underline{0.0338}}
& 0.2383
& \textcolor{red}{\textbf{0.0336}}
& 0.0348 \\

\bottomrule
\end{tabular}
\end{minipage}

\end{table*}

\begin{table*}[t]
\caption{Standard deviations of the CRPS results reported in
Table~\ref{tab:crps_point_block}.}
\label{tab:std_crps_point_block}
\vspace{-1ex}

\begin{minipage}[t]{0.66\textwidth}
\centering
\caption*{(a) Point missing}
\setlength{\tabcolsep}{4pt}
\renewcommand{\arraystretch}{0.95}
\begin{tabular}{l|l|cccc}
\toprule
\multicolumn{2}{c|}{Models}
& \textbf{\model{}}
& GP-VAE
& CSDI
& FGTI \\
\multicolumn{2}{c|}{Metric}
& CRPS & CRPS & CRPS & CRPS \\
\midrule

\multirow{5}{*}{\rotatebox{90}{ETTh1}}
& 0.2 & 0.0012 & 0.0073 & 0.0023 & 0.0021 \\
& 0.4 & 0.0017 & 0.0050 & 0.0046 & 0.0031 \\
& 0.6 & 0.0017 & 0.0056 & 0.0045 & 0.0044 \\
& 0.8 & 0.0064 & 0.0098 & 0.0103 & 0.0091 \\
\cmidrule(lr){2-6}
& Avg & 0.0027 & 0.0069 & 0.0054 & 0.0047 \\
\midrule

\multirow{5}{*}{\rotatebox{90}{ETTh2}}
& 0.2 & 0.0022 & 0.0261 & 0.0037 & 0.0030 \\
& 0.4 & 0.0019 & 0.0269 & 0.0044 & 0.0019 \\
& 0.6 & 0.0027 & 0.0264 & 0.0061 & 0.0010 \\
& 0.8 & 0.0050 & 0.0134 & 0.0116 & 0.0229 \\
\cmidrule(lr){2-6}
& Avg & 0.0030 & 0.0232 & 0.0065 & 0.0072 \\
\midrule

\multirow{5}{*}{\rotatebox{90}{Exchange}}
& 0.2 & 0.0004 & 0.0253 & 0.0652 & 0.0011 \\
& 0.4 & 0.0005 & 0.0220 & 0.0798 & 0.0025 \\
& 0.6 & 0.0003 & 0.0220 & 0.0944 & 0.0043 \\
& 0.8 & 0.0002 & 0.0298 & 0.1247 & 0.0182 \\
\cmidrule(lr){2-6}
& Avg & 0.0004 & 0.0248 & 0.0910 & 0.0065 \\
\midrule

\multirow{5}{*}{\rotatebox{90}{Illness}}
& 0.2 & 0.0038 & 0.0058 & 0.0099 & 0.0079 \\
& 0.4 & 0.0028 & 0.0250 & 0.0077 & 0.0128 \\
& 0.6 & 0.0030 & 0.0206 & 0.0094 & 0.0171 \\
& 0.8 & 0.0149 & 0.0195 & 0.0155 & 0.0380 \\
\cmidrule(lr){2-6}
& Avg & 0.0061 & 0.0177 & 0.0106 & 0.0190 \\
\midrule

\multirow{5}{*}{\rotatebox{90}{Weather}}
& 0.2 & 0.0005 & 0.0165 & 0.0008 & 0.0032 \\
& 0.4 & 0.0008 & 0.0152 & 0.0012 & 0.0039 \\
& 0.6 & 0.0020 & 0.0186 & 0.0017 & 0.0043 \\
& 0.8 & 0.0034 & 0.0421 & 0.0020 & 0.0066 \\
\cmidrule(lr){2-6}
& Avg & 0.0017 & 0.0231 & 0.0014 & 0.0045 \\
\bottomrule
\end{tabular}
\end{minipage}
\hfill
\begin{minipage}[t]{0.30\textwidth}
\centering
\caption*{(b) Block missing}
\setlength{\tabcolsep}{3pt}
\renewcommand{\arraystretch}{1.0}
\begin{tabular}{lcccc}
\toprule
Dataset & \textbf{\model{}} & GP-VAE & CSDI & FGTI \\
\midrule
ETTh1    & 0.0015 & 0.0080 & 0.0038 & 0.0027 \\
ETTh2    & 0.0021 & 0.0326 & 0.0107 & 0.0185 \\
Exchange & 0.0070 & 0.0188 & 0.1331 & 0.0320 \\
Illness & 0.0546 & 0.0783 & 0.1111 & 0.0969 \\
Weather & 0.0032 & 0.0194 & 0.0029 & 0.0058 \\
\bottomrule
\end{tabular}
\end{minipage}

\end{table*}

\FloatBarrier
\section{Additional Qualitative Results}
\label{app:additional_qualitative_results}
We provide additional qualitative imputation visualizations under varying point wise missing ratios (20\%, 40\%, 60\%, and 80\%) and block missingness for ETTh2. These figures illustrate median predictions and 90\% predictive intervals, highlighting reconstruction accuracy and uncertainty calibration.

\FloatBarrier
\begin{figure}[h]
    \centering
    \includegraphics[width=0.95\columnwidth]{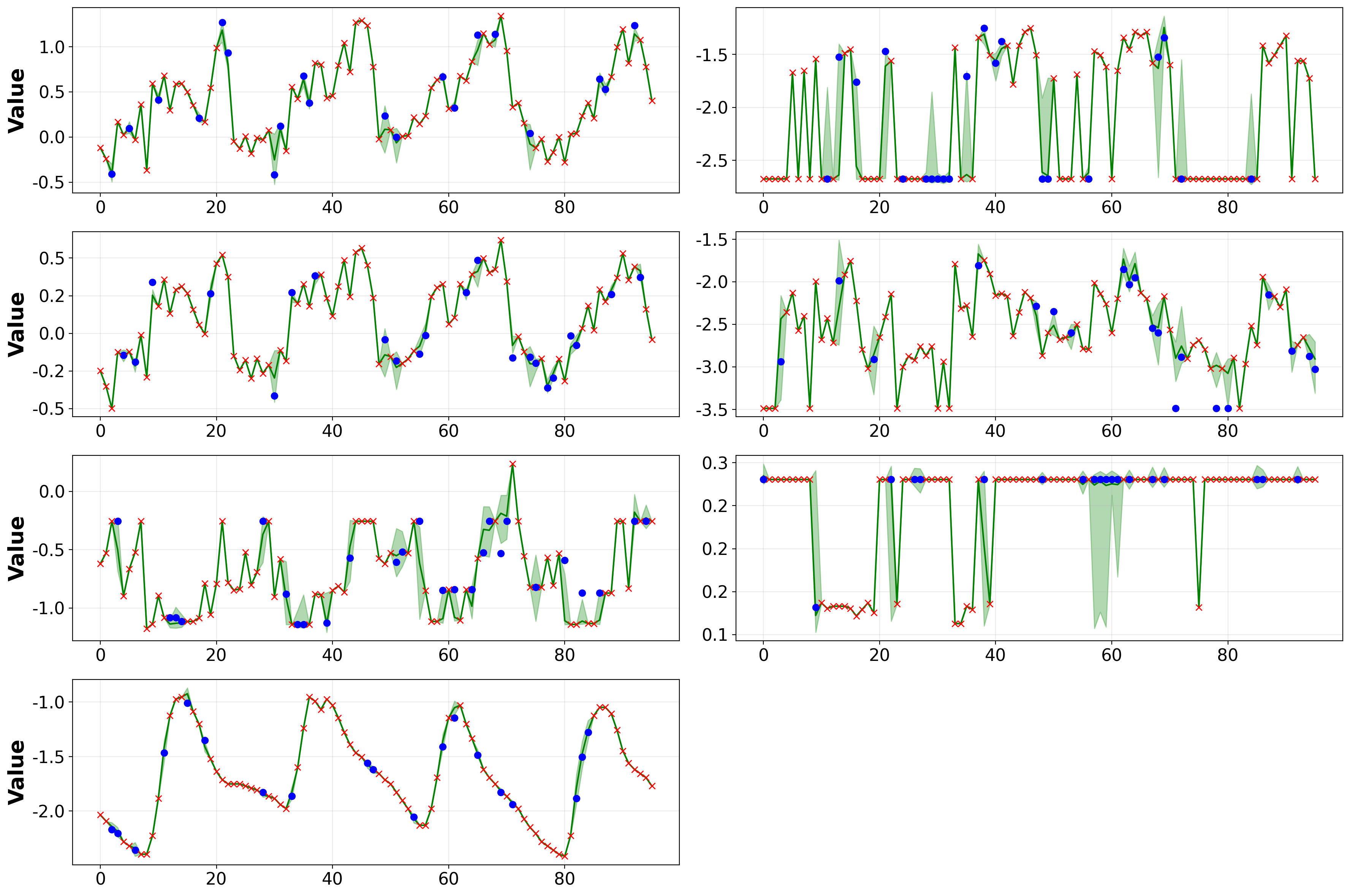}
    \caption{Visualization of probabilistic imputation results on ETTh2 (20\% missingness). The results are for a time series sample with all 7 features. The median and 90\% predictive intervals are shown in green, observed points in red and targets in blue.}
    \label{fig:all_feat1}

\end{figure}

\begin{figure}[h]
    \centering
    \includegraphics[width=0.95\columnwidth]{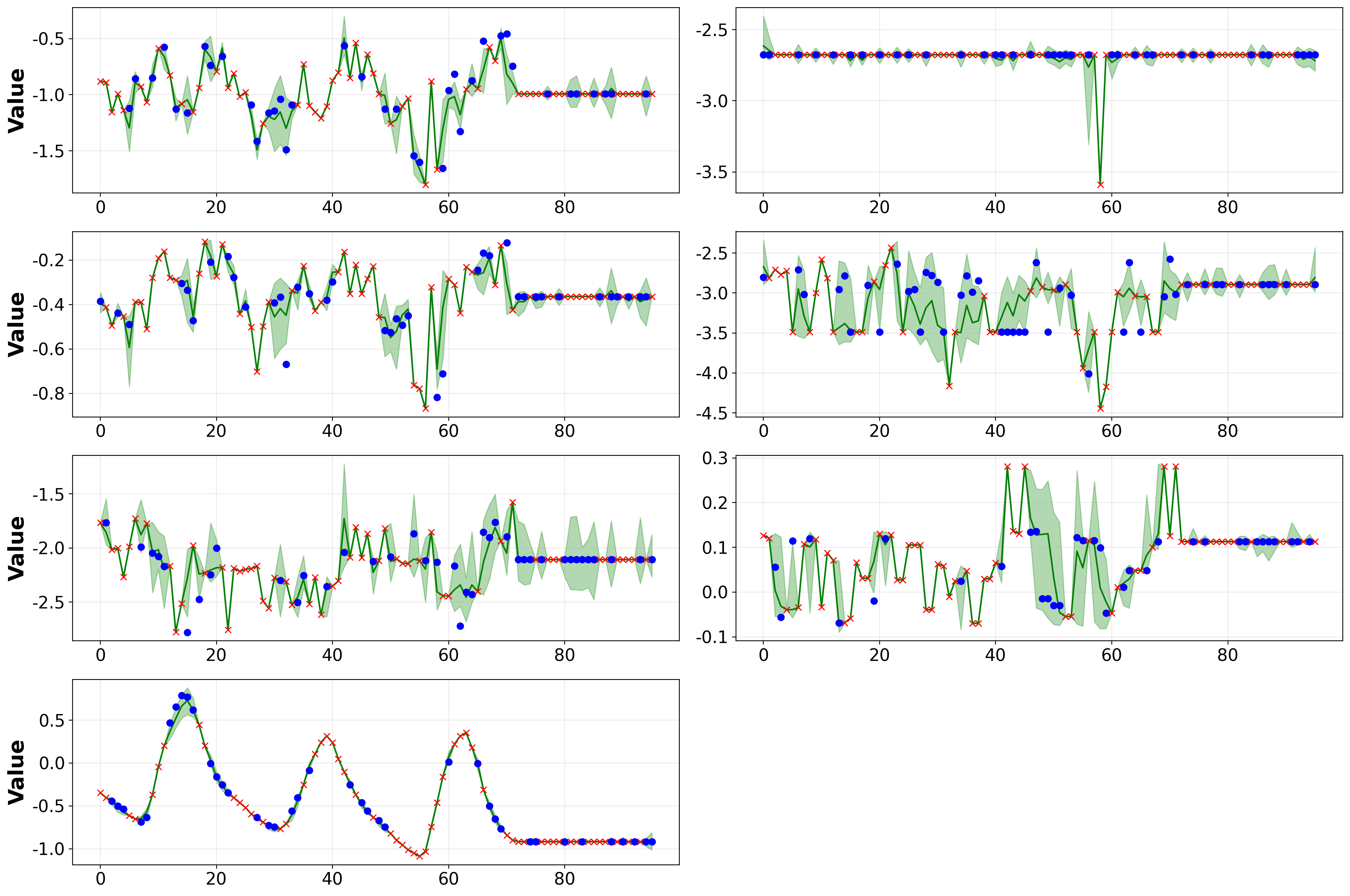}
    \caption{Visualization of probabilistic imputation results on ETTh2 (40\% missingness). The results are for a time series sample with all 7 features. The median and 90\% predictive intervals are shown in green, observed points in red and targets in blue.}
    \label{fig:all_feat4}

\end{figure}

\begin{figure}[h]
    \centering
    \includegraphics[width=0.95\columnwidth]{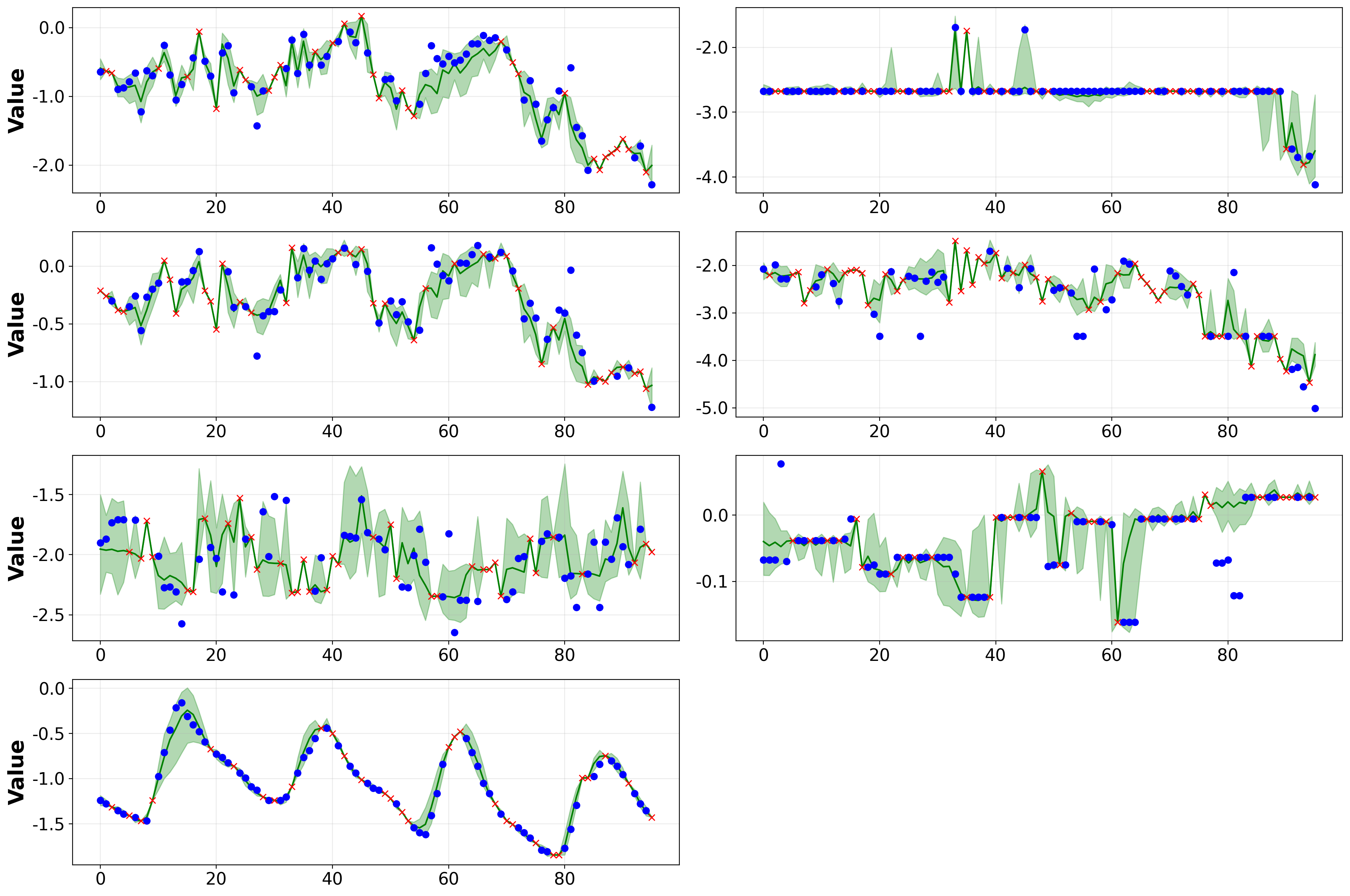}
    \caption{Visualization of probabilistic imputation results on ETTh2 (60\% missingness). The results are for a time series sample with all 7 features. The median and 90\% predictive intervals are shown in green, observed points in red and targets in blue.}
    \label{fig:all_feat6}

\end{figure}

\begin{figure}[h]
    \centering
    \includegraphics[width=0.95\columnwidth]{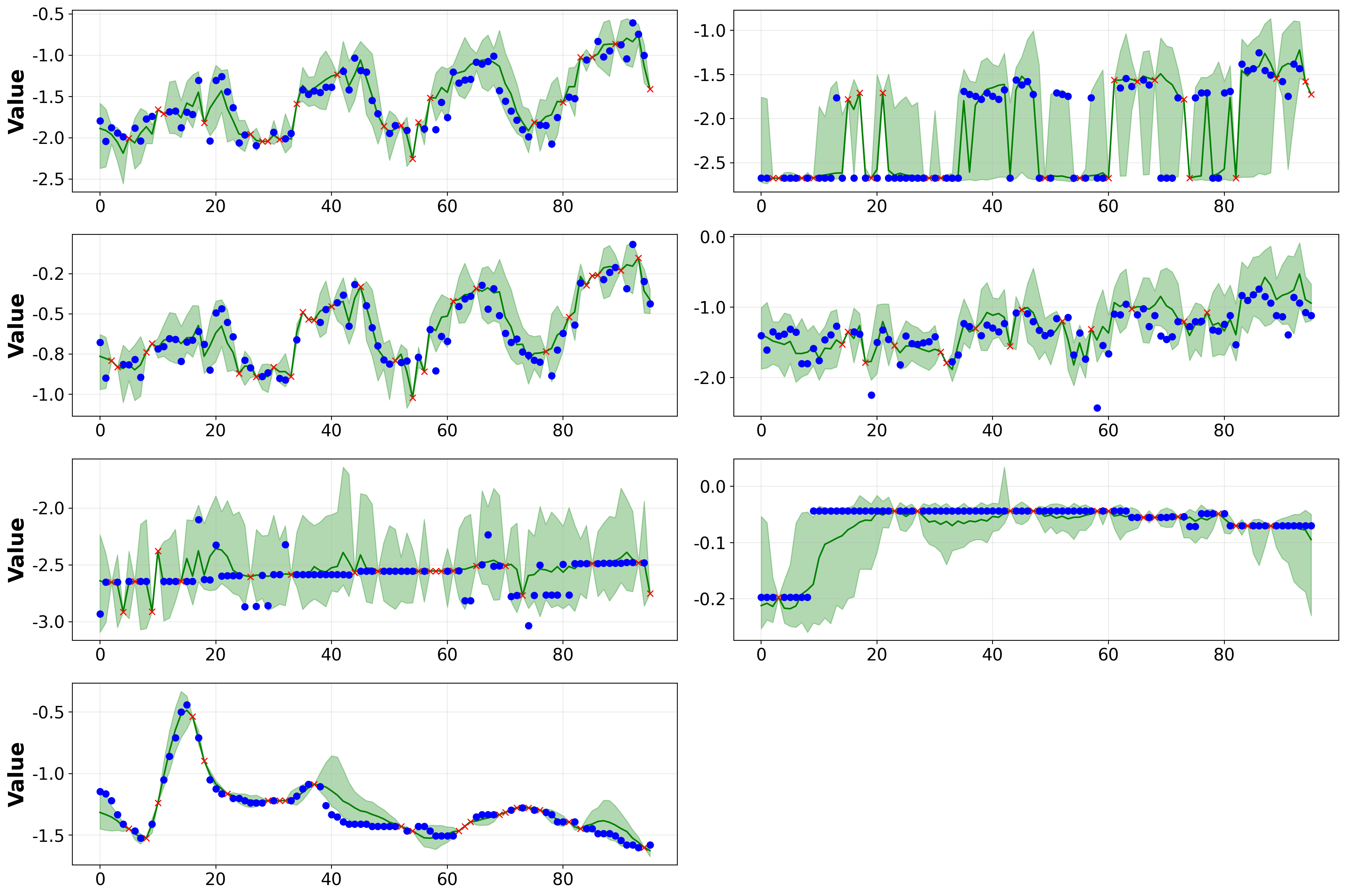}
    \caption{Visualization of probabilistic imputation results on ETTh2 (80\% missingness). The results are for a time series sample with all 7 features. The median and 90\% predictive intervals are shown in green, observed points in red and targets in blue.}
    \label{fig:all_feat8}

\end{figure}

\begin{figure}[h]
    \centering
    \includegraphics[width=0.95\columnwidth]{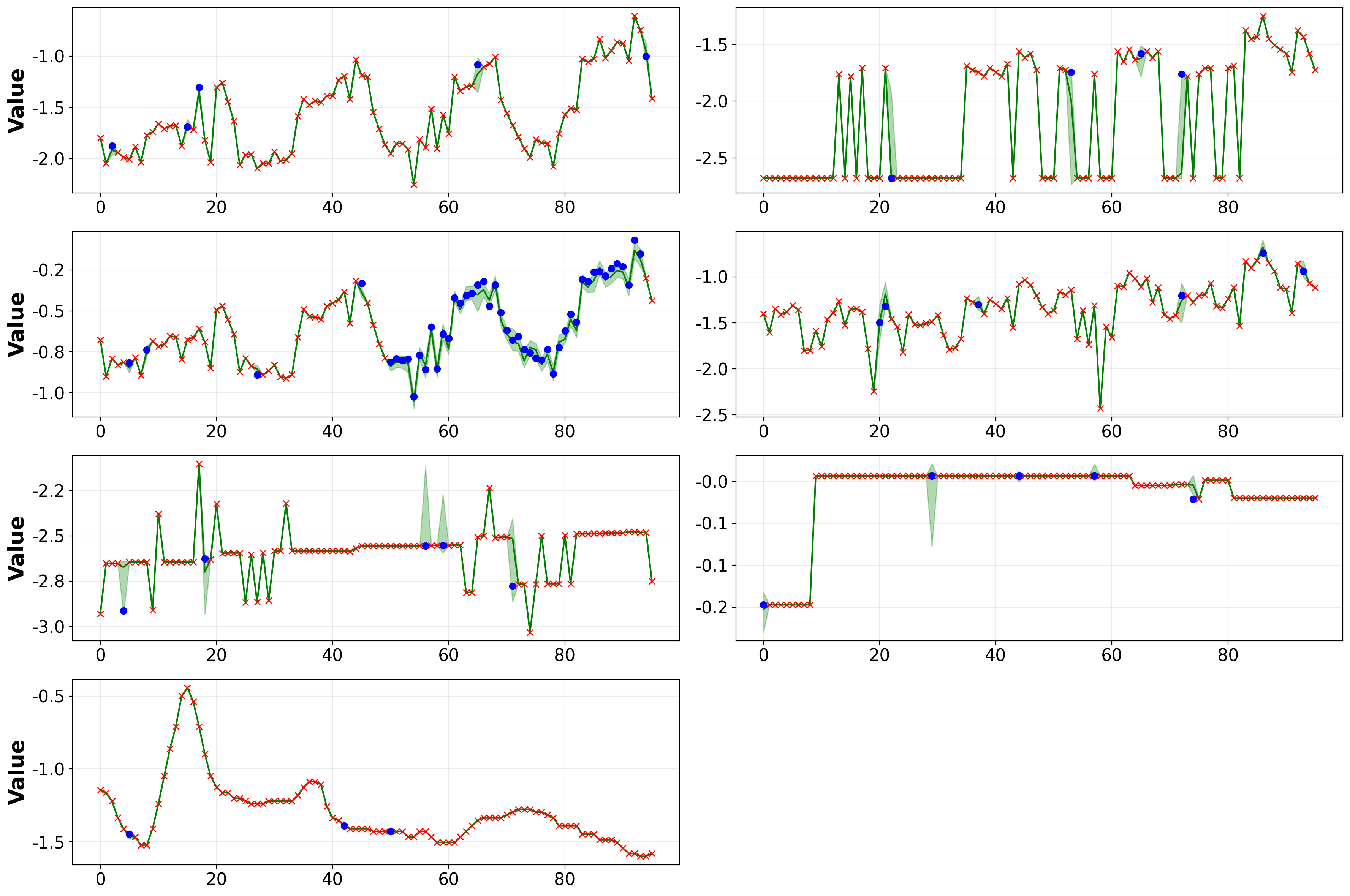}
    \caption{Visualization of probabilistic imputation results on ETTh2 (block missingness). The results are for a time series sample with all 7 features. The median and 90\% predictive intervals are shown in green, observed points in red and targets in blue.}
    \label{fig:all_feat_block}

\end{figure}

\FloatBarrier